# Response Transformation and Profit Decomposition for Revenue Uplift Modeling

Robin M. Gubela[1], Stefan Lessmann[1], Szymon Jaroszewicz[2]

[1] School of Business and Economics, Humboldt-University of Berlin, Unter den Linden 6, 10099 Berlin, Germany

[2] Institute of Computer Science, Polish Academy of Sciences, ul. Jana Kazimierza 5, 01-248 Warsaw, Poland

**Abstract**

Uplift models support decision-making in marketing campaign planning. Estimating the causal effect of a marketing treatment, an uplift model facilitates targeting communication to responsive customers and an efficient allocation of marketing budgets. Research into uplift models focuses on conversion models to maximize incremental sales. The paper introduces uplift modeling strategies for maximizing incremental revenues. If customers differ in their spending behavior, revenue maximization is a more plausible business objective compared to maximizing conversions. The proposed methodology entails a transformation of the prediction target, customer-level revenues, that facilitates implementing a causal uplift model using standard machine learning algorithms. The distribution of campaign revenues is typically zero-inflated because of many non-buyers. Remedies to this modeling challenge are incorporated in the proposed revenue uplift strategies in the form of two-stage models. Empirical experiments using real-world e-commerce data confirms the merits of the proposed revenue uplift strategy over relevant alternatives including uplift models for conversion and recently developed causal machine learning algorithms. To quantify the degree to which improved targeting decisions raise return on marketing, the paper develops a decomposition of campaign profit. Applying the decomposition to a digital coupon targeting campaign, the paper provides evidence that revenue uplift modeling as well as causal machine learning can improve campaign profit substantially.

# 1  Introduction

Predictive models are a popular tool to support decision-making in marketing. Exemplary use cases include forecasting usage frequencies in social media (e.g., Ballings & Van den Poel, 2015), response toward advertisement (e.g., Goldfarb & Tucker, 2011), and customer churn probabilities (e.g., Verbeke et al., 2012). We distinguish two categories of marketing prediction models, response and uplift models. Response models predict customer behavior in general. The term was coined in the direct marketing literature where the modeling goal is often to predict how a customer will react to a marketing stimulus (e.g., Baesens et al., 2002). Uplift models also consider a direct marketing setting but estimate the differential change in response behavior due to the marketing activity (Devriendt et al., 2018). This way, an uplift model accounts for the causal link between the action and customer response. Causality is crucial to measure the true impact of a marketing campaign, maximize campaign profit, and allocate scarce marketing resources efficiently (e.g., Lo & Pachamanova, 2015). The paper proposes novel methodologies for uplift modeling to support targeting decisions in marketing campaign planning.

We consider promotional campaigns that aim at maximizing sales revenues in e-commerce through issuing digital coupons (e.g., Reimers & Xie, 2019). The prevalence of coupons suggests that coupon targeting is a relevant task. For example, US retailers distributed 256.5 billion coupons for consumer packaged goods in 2018, and consumers redeemed over 1.7 billion coupons with a combined face value of $2.7 billion (NCH Marketing Services, 2019). Issuing a coupon equates to a price reduction. In this regard, coupon allocation re-emphasizes the cruciality of using a causal targeting model. Offering a coupon to a customer who would buy at the ordinary price wastes marketing resources and decreases sales margin. Therefore, a targeting model must identify those customers who have no intention to buy but can be persuaded to buy through a discount. This is the aim of an uplift model, while a response model could only predict the buying propensity of customers, which is less useful for targeting (e.g., Ascarza, 2018).

Prior research on uplift models focuses on applications with a binary prediction target. The target might be a purchase incident. An uplift model could then estimate the change in customers' purchase probabilities due to a marketing activity (e.g., a coupon). We call such models conversion models since they capture whether the marketing activity has altered customer behavior, where behavior refers to a binary event such as redeeming a coupon, clicking a link, or buying a product (e.g., Devriendt et al., 2018). Targeting campaigns using a conversion uplift model implies a sales maximization objective in that conversion uplift, by definition, captures incremental customer actions. Due to heterogeneity in customer spending (e.g., Schröder & Hruschka, 2017), maximizing incremental conversions and maximizing incremental profits do not necessarily coincide. Considering our application setting, the ideal customer to offer a coupon would be a person who has no buying intention initially, is successfully persuaded to buy by the coupon, and then spends a large amount. Conversely, converting a customer who spends a small amount is less valuable and might have a negative profit impact if the treatment costs associated with coupon provision exceed incremental revenues. The overarching goal of the paper is to introduce revenue uplift modeling. Similar to a conversion model, which strives to maximize incremental sales, a revenue uplift model aims at maximizes incremental revenues. In applications where customers exhibit heterogeneity in spending/value, such as coupon targeting where customers differ in the value of their market baskets, revenue uplift modeling represents a more direct approach to maximize campaign profit. It facilitates targeting a campaign to those customers who generate the largest incremental revenue. The paper develops approaches to implement revenue uplift using supervised machine learning (SML) algorithms and tests their effectiveness through empirical experimentation in the scope of e-coupon targeting.

A revenue uplift model estimates the incremental (i.e. due to a coupon) sales revenue on an individual customer level. The continuous prediction target prohibits the use of existing uplift modeling approaches for conversion, where the target variable is binary. A first contribution of the paper is the extension of the binary target variable transformation (Lai et al., 2006), a competitive method to develop conversion uplift models (Devriendt et al., 2018; Kane et al., 2014), to accommodate a continuous target variable. We propose two transformations that facilitate developing a revenue uplift model using SML algorithms for regression or classification. This way, marketers have full flexibility to choose their preferred learning algorithm, and use their preferred SML software package, to estimate a causal revenue uplift model.

In addition to a change in the scaling level of the prediction target, a second methodological challenge in revenue uplift modeling concerns the revenue distribution, which is typically skewed. Considering our application setting, most visitors to an online shop do not buy and generate zero revenues. Likewise, coupon redemption rates in grocery retailing are only 0.6% in the US (NCH Marketing Services). More generally, the issue of low response rates has a long tradition in direct marketing literature (e.g., Magliozzi & Berger, 1993). It is also known that a skewed distribution of the target variable impedes regression modeling, for example in the context of credit risk models for loss given default prediction (Yao et al., 2017). While remedies to skewed response distributions in the form of zero-inflated regression and hurdle models are well-established in other domains, we are not aware of previous work in uplift modeling that addresses skewed distributions of the target variable through specific modeling solutions. A second contribution of the paper is that it introduces and tests a two-stage modeling framework for revenue uplift modeling.

A third contribution concerns the evaluation of (revenue) uplift models. We propose a novel profit decomposition for marketing campaigns to measure the incremental profit impact of an uplift model. The campaign profit measure is easily interpretable and bridges the gap between model-based forecasts and financial business goals. Incorporating a comprehensive cost model, the profit measure supports different types of marketing campaigns including, but not limited, to coupon targeting and generalizes previous decompositions of campaign profit (e.g., Lessmann et al., 2019; Verbeke et al., 2012) to uplift modeling.

Last, capturing the impact of a marketing activity on customer behavior, uplift modeling is closely related to the literature on causal inference and treatment effects (e.g., Imbens & Rubin, 2015). Using the terminology of that literature, uplift modeling corresponds to estimating conditional average treatment effects (CATE), where the conditioning is based on customer characteristics (Knaus et al., 2018). Aiming at comparing customers, an uplift model emphasizes individual-level effects, which correspond to the finest level of conditioning. Since the fundamental problem of causal inference (Holland, 1986) renders individualized effects unobservable, a more common term is that of on an individualized average treatment effect; sometimes also called personalized treatment effect (Guelman et al., 2015a). In principle, any approach for CATE estimation facilitates uplift modeling. However, we note a subtle difference between an uplift model and a CATE estimator, which arises in a campaign targeting context. Targeting a marketing activity aims at an accurate ranking of customers. This way, a marketer can approach only top-ranked customers, whereby the ranking criterion is the CATE. First estimating individualized treatment effects and then sorting customers accordingly is a viable approach toward uplift modeling. However, a model that gives biased estimates of the CATE – or foregoes its estimation altogether – can still be a valuable uplift model if the relative order of customers in the ranked list is accurate. Uplift modeling strategies such as the class variable transformation, which display appealing results in several benchmarks (Devriendt et al., 2018; Gubela et al., 2019; Kane et al., 2014), and their extensions as proposed in this paper, belong to the latter category. In view of recent advancements in the literature on treatment effects and the development of several highly recognized methods such as causal forests (Athey et al., 2019), causal boosting (Powers et al., 2018), causal BART (Hill, 2011), or the X-learner of Künzel et al. (2019), it is interesting to examine whether these CATE estimators are a suitable vehicle for revenue uplift modeling. A fourth empirical contribution of the paper is thus that it assesses the merit of recently developed causal machine learning algorithms for revenue uplift modeling and compares their effectiveness to the proposed revenue uplift strategies.

## 2 Background and Related Work

Marketing decision models for differential response analysis (Radcliffe & Surry, 1999) or incremental value modeling (Chickering & Heckerman, 2000) have a long tradition. A seminal paper by Lo (2002) coined the term true lift, which has later been revised to uplift (Radcliffe & Surry, 2011). Most previous studies explain in detail how an uplift model differs from a classical response model and estimates causal effects (e.g., Lo & Pachamanova, 2015). However, especially earlier papers do not emphasize connections between uplift and the body of literature on treatment effects. Our review aims at providing a comprehensive overview of previous work in different fields and substantiating how the paper adds to the literature.

We consider previous studies on uplift modeling and machine learning methods for CATE estimation. The focus on machine learning is suitable because marketing data sets are typically large and high-dimensional. Recently developed causal machine learning algorithms highlight such data characteristics as design goals and innovation over more traditional methods for treatment effect estimation (Knaus et al., 2018). This suggests that machine learning-based models are well prepared for marketing applications. Overall, our review of prior work identifies 38 previous studies that have contributed novel methodology to the field of uplift analytics. We organize these studies along the stages of the KDD process model (Fayyad et al., 1996) to identify their methodological contributions in Table 1. The KDD process identifies modeling steps that occur in every data analytics initiative. Therefore, it provides a suitable framework to systematize the interdisciplinary previous work on uplift modeling. Note that Table 1 excludes recent benchmarking studies related to (conversion) uplift and CATE (Devriendt et al., 2018; Diemert et al., 2018; Gubela et al., 2019; Knaus et al., 2018) since these papers contribute empirical insight but do not provide novel methodology.

To capture a causal link between a marketing activity and customer behavior, uplift models require data from two groups, the treatment and the control group (Devriendt et al., 2018). Such data is gathered through randomized trials in previous work; often in the form of A/B tests in e-commerce settings. The existence of two disjoint sets of data represents a major difference to conventional applications of SML. KDD stages need to address this difference. Some studies have explicitly considered issues related to data source selection (e.g., Diemert et al., 2018). Likewise, handling treatment and control group data requires adjustments of data preparation tasks as examined by a few studies (e.g., Hansen & Bowers, 2008; Hua, 2016) and necessitates novel measures to assess model quality (e.g., Nassif et al., 2013).

However, Table 1 clarifies that the focus in most previous studies was to develop novel algorithms for the estimation of uplift models (i.e., CATE). An extension of tree-based algorithms by new splitting criteria has been particularly popular. Their partitioning mechanism makes decision trees suitable to process different subsets of data (e.g., treatment/control group observations). Hence, early uplift papers (e.g., Hansotia & Rukstales, 2002b) and recent papers in causal machine learning (e.g., Athey et al., 2019; Powers et al., 2018) consider a tree learning framework and propose splitting criteria for CATE estimation. However, other popular SML algorithms such as support vector machines (e.g., Kuusisto et al., 2014) or neural networks (e.g., Shalit et al., 2017) have also been extended to uplift settings.

The focus of this paper is on the data transformation step of the KDD process. This step is interesting because a clever transformation of the input (i.e., covariate) or output (i.e., target variable) space of a modeling data set facilitates the development of an uplift model using standard SML algorithms. From a practitioner's point of view, avoiding the development of a tailor-made learning algorithm has some advantages. It facilitates capitalizing on the large set of available SML algorithms and their scalable implementations in software packages (e.g., Kochura et al., 2017). Tailor-made causal machine learning algorithms such as causal forests (Athey et al., 2019), boosting (Powers et al., 2018) or BART (Hill, 2011) show excellent results but do not offer algorithmic flexibility. A tree learning framework is not the best choice for any data set. Being able to choose any learning algorithm for uplift estimation is preferable. The two-stage revenue uplift models, which we propose below to address zero-inflated revenue distributions, are a good example for the merit of (re-)using SML algorithms for uplift estimation. While the development of a corresponding approach, for example, a zero-inflated causal neural network, is an interesting avenue for research, marketing professionals will appreciate the far simpler solution to devise a modeling solution based on existing, proven methodology like SML. Furthermore, given that causal machine learning is yet a young and emerging field, frameworks for high-performance computing such as Apache Spark do not yet support corresponding techniques. Also, available implementations of causal machine learning methods may be optimized for efficiency. Processing the huge amounts of data that routinely occur in corporate marketing may be challenging with existing causal machine learning algorithms. Finally, the process of developing a targeting model utilizing uplift transformations is almost the same as in conventional direct marketing. Providing a gradual transition of established targeting practices to a causal modeling framework and leveraging existing technology, we suggest that uplift transformation approaches are suitable to support campaign planning in corporate environments.

Table 1 displays that previous literature on uplift transformation is relatively sparse. Lo (2002) and Tian et al. (2014) use (logistic) regression models to estimate uplift. To capture causality, they augment their models with interaction terms of a binary treatment-control-group indicator variable with other covariates. We formally introduce this approach in Section 3. Lai et al. (2006) propose an alternative strategy, Lai's weighted uplift method (LWUM), which was further refined by Kane et al. (2014). The approach involves a modification of the target variable. Both regimes, covariate and response transformation, facilitate the use of SML algorithms to model uplift. The above studies have considered binary target variables (e.g., conversion uplift). The focus of this paper is revenue uplift, which involves processing a continuous target variable. Processing continuous target variables in a covariate transformation framework is straightforward and has been considered by Tian et al. (2014). As will become clear in Section 3, a corresponding extension to support continuous target variables involves major adjustments for the response variable transformation approach. Developing response variable transformations for continuous target variables and revenue uplift modeling is thus the focus of this.

Uplift data mining techniques based on decision trees are an alternative to process continuous response variables. In the empirical part of the paper, we consider tree-based models for uplift and CATE estimation as a benchmark to set the performance of the proposed uplift transformation strategies into context. Given that none of the tree-based techniques has been employed for modeling revenue uplift in e-commerce, the evaluation also broadens the scope of empirical results for causal machine learning methods.

**Table 1: Methodological innovations in prior work on uplift modeling and machine learning for CATE**

| Publication | Research Focus in terms of KDD Process | | | | | | |
|---|---|---|---|---|---|---|---|
| | Data Selection | Data Pre-Processing | Uplift Transformation | | | Uplift / CATE Estimation | Evaluation |
| | | | Conversion Response Transformation | Revenue Response Transformation | Covariates Transformation | | |
| Athey and Imbens (2016) | | | | | | Causal Tree | |
| Athey et al. (2019) | | | | | | Causal Forest | |
| Cai et al. (2011) | | | | | | Two-Step Estimation Procedure | |
| Chickering and Heckerman (2000) | | | | | | Uplift Tree with Post-Processing Procedure | |
| Diemert et al. (2018) | x | | | | | - | |
| Guelman et al. (2015a) | | | | | | Causal Conditional Inference Tree/Forest | |
| Guelman et al. (2015b) | | | | | | Uplift Random Forests | |
| Gutierrez and Gérardy (2017) | | | | | | - | x |
| Hahn et al. (2019) | | | | | | Causal Bayesian Regression Trees | |
| Hansen and Bowers (2008) | | x | | | | - | |
| Hansotia and Rukstales (2002a) | | | | | | Incremental Response Tree | |
| Hansotia and Rukstales (2002b) | | | | | | Uplift Tree with the $\Delta\Delta P$ splitting criterion | |
| Hill (2011) | | | | | | Causal BART | |
| Imai and Ratkovic (2013) | | | | | | Uplift Support Vector Machine | |
| Jaroszewicz and Rzepakowski (2014) | | | | | | Uplift k-Nearest Neighbors | |
| Kane et al. (2014) | x | | | | | - | |
| Kuusisto et al. (2014) | | | | | | Uplift Support Vector Machine | |
| Künzel et al. (2019) | | | | | | X-Learner | |
| Lai et al. (2006) | | | x | | | - | |
| Lechner (2019) | | | | | | Modified Causal Forests | |
| Lo (2002) | | | | | x | - | |
| Lo and Pachamanova (2015) | | | | | | Multiple Treatments Logistic Regression | |
| Nassif et al. (2013) | | | | | | - | x |
| Oprescu et al. (2018) | | | | | | Orthogonal Causal Random Forest | |
| Powers et al. (2018) | | | | | | Causal boosting | |
| Radcliffe and Surry (1999) | | | | | | Uplift Trees | |
| Radcliffe and Surry (2011) | | | | | | - | x |
| Rzepakowski and Jaroszewicz (2012a) | | | | | | Multiple Treatments Uplift Trees | |
| Rzepakowski and Jaroszewicz (2012b) | | | | | | Information Theory-Based Uplift Trees | |
| Rudaś and Jaroszewicz (2018) | | | | | x | - | |
| Shaar et al. (2016) | | | | | | Pessimistic Uplift | |
| Shalit et al. (2017) | | | | | | Causal Artificial Neural Network | |
| Sołtys et al. (2015) | | | | | | Uplift Ensemble Methods | |
| Su et al. (2012) | | | | | | Uplift k-Nearest Neighbors | |
| Taddy et al. (2016) | | | | | | Causal Bayesian Forests | |
| Tian et al. (2014) | | | | | x | - | |
| Yamane et al. (2018) | | | | | | Separate Label Uplift Modeling | |
| *This study* | | | | x | | - | x |

Note: Only the primary contribution of a paper has been considered for the systematization of related literature.

# 3 Revenue Uplift Modeling

In the following, we review uplift modeling fundamentals and introduce our notation. Thereafter, we elaborate on target variable and covariate transformations for revenue uplift modeling. Finally, we introduce our two-stage models to address zero-inflated revenue distributions.

## 3.1 Uplift Modeling Fundamentals

Let $X_i = (x_{i1}, \ldots, x_{in}) \in \mathbb{R}^n$ be a vector of size n that represents individual customers. The elements of $X_i$ capture customer characteristics such as demographic or behavioral attributes. Let $Y_{i,c} \in \{0,1\}$ be the binary response in a conversion model, with $Y_{i,c} = 1$ indicating, for example, purchase of a product. The index $i = 1, \ldots, N$ refers to individual customers. Throughout the paper, we use the subscripts $c$ and $r$ to distinguish between a conversion and revenue setting. Further, let $T_i \in \{0,1\}$ be an indicator of a customer's group affiliation, with $T_i = 0$ and $T_i = 1$ indicating control and treatment group customers, respectively. As purchase probabilities equal response expectations, or more formally, $P(Y = 1) = E(Y)$, let $E(Y_i|X_i, T_i = 1)$ and $E(Y_i|X_i, T_i = 0)$ denote the group-specific response expectations. This notation facilitates a formal delineation of response and uplift models. Response models aim at predicting model lift:

$$Lift_{i,c}^{Response} = E(Y_{i,c}|X_i, T_i = 1)/E(Y_{i,c}) = E(Y_{i,c}|X_i, T_i)/E(Y_{i,c}). \qquad (1)$$

The predictions are then used to select customers for which this quantity is the largest. Given that the prior expectation of responding, $E(Y_{i,c})$, is a characteristic of the data, the primary task of a response model is to estimate the posterior expectations $E(Y_{i,c}|X_i, T_i = 1)$. Uplift models predict uplift as the change in behavior resulting from a promotional marketing activity, which is equivalent to a change in the posterior expectation with and without treatment and thus the causal effect of the treatment (e.g., Imbens & Rubin, 2015):

$$Uplift_{i,c}^{Indirect} = E(Y_{i,c}|X_i, T_i = 1) - E(Y_{i,c}|X_i, T_i = 0). \qquad (2)$$

Estimated class expectations, also called lift and uplift scores, respectively, facilitate a ranking of customers and targeting top-ranked customers with a campaign (e.g., Devriendt et al., 2018).

Equation (2) illustrates a possible strategy to develop a conversion uplift model through estimating two classification models from treatment and control group observations, respectively using some SML algorithm. Subtraction of the expectations on the two mirrors the idea of uplift modeling, the objective of which is to maximize the number of treatment responders while minimizing control responders based on model predictions. This strategy is known as indirect uplift modeling (e.g., Kane et al., 2014) as it predicts uplift using two independent classifiers. The indirect uplift approach suffers from conceptual drawbacks that often lead to poor model performance Hansotia and Rukstales (2002b). This has motivated the development of alternative conversion uplift modeling strategies (see Table 1).

Our research focusses on revenue uplift modeling. We first introduce the most fundamental revenue strategies, which are revenue response modeling and the indirect revenue uplift approach. In contrast to their analogs in conversion modeling, these differ in that they take a (continuous) revenue response $Y_{i,r} \in \mathbb{R}$ instead of a (binary) conversion target $Y_{i,c}$ into account. In formal terms, revenue response models predict response by

$$Lift_{i,r}^{Response} = E(Y_{i,r}|X_i, T_i = 1), \qquad (3)$$

whereas indirect revenue uplift models predict uplift by

$$Uplift_{i,r}^{Indirect} = E(Y_{i,r}|X_i, T_i = 1) - E(Y_{i,r}|X_i, T_i = 0). \tag{4}$$

Irrespective of a modeling strategy, targeting models are developed from an uplift modeling data set of the form $\{Y_{i,r}, X_i, T_i\}_{i=1}^N$. In marketing, such data stems from a previous campaign or may be gathered using A/B tests. More generally, randomized trials are a typical way to obtain modeling data. In our setting, $Y_{i,r}$ represents the total purchase amount of customer i, $X_i$ includes characteristics of this customer, and $T_i$ indicates whether the customer has received a marketing stimulus. We assume conditional independence, that is that values of $T_i$ have been generated at random independent from $X_i$'s. This assumption is critical to mitigating bias of model estimates (Imbens & Rubin, 2015).

## 3.2 Revenue Uplift Transformations

### 3.2.1 Revenue Response Transformation

We propose two response transformation approaches for revenue modeling, which we call continuous response variable transformation with weightings (CRVTW), and revenue discretization transformation (RDT). The approaches aim to transfer the treatment/control group information from the data input space (i.e., independent variables) to its output space (i.e., the dependent variable).

CRVTW transforms the response variable in such a way that it captures critical information from treatment and control groups. More specifically, we define $z_{i,rw} \in \mathbb{R}$ as follows:

$$z_{i,rw} = \begin{cases} +\frac{1}{q^T} Y_{i,r} & \text{if } T_i = 1 \wedge Y_{i,r} > 0 \\ -\frac{1}{q^C} Y_{i,r} & \text{if } T_i = 0 \wedge Y_{i,r} > 0 \\ 0 & \text{otherwise.} \end{cases} \tag{5}$$

with $q^T = n^T/n$ and $q^C = n^C/n$ as the fractions of treatment or control group customers from the whole population. The weightings facilitate unbiased estimation. For formal proofs on statistical properties, we refer to Rudaś and Jaroszewicz (2018). The transformed response variable is positive if the customer responded positively to the incentive and provided a purchase value of $Y_{i,r}$. For purchasers of the control group, $Y_{i,r}$ corresponds to the negative purchase volume. The remaining cases include treatment and control group customers without purchase, and we define $z_{i,rw}$ to be zero in these cases.

CRVTW draws inspiration from Lai's method for conversion uplift modeling (Lai et al., 2006), which, in the light of recent benchmarking results (e.g., Devriendt et al., 2018; Kane et al., 2014), can be considered a promising modeling strategy for conversion uplift. Equation (5) extends this approach to continuous response variables. A standard regression model suffices to model $z_{i,rw}$, which itself possesses all relevant information to capture uplift. Using some regression method to learn a functional relationship between $z_{i,rw}$ and $X_i$ provides a model that distinguishes purchasing customers from the treatment and control groups while also capturing the dependency between covariates and purchase amounts. We assume that targeting customers according to the predictions of such uplift model (i.e., soliciting customers in descending order of scores from prediction of $z_{i,rw}$) leads to a target group of receptive, or persuadable, customers with high willingness to spend. Following the same logic, we assume scores from predicting $z_{i,rw}$ for customers who are ready to buy without solicitation and thus not in need of a marketing investment to be low (i.e., negative), resulting in them being ignored in targeted marketing activities. Therefore, CRVTW represents an uplift transformation and facilitates the development of a revenue uplift model through estimating (6) using some SML algorithm for regression:

$$Uplift_{i,r}^{CRVTW} = E(z_{i,rw}|X_i). \qquad (6)$$

The second transformation we propose, RDT, is based on Bodapati and Gupta (2004) who recommend a discretization of continuous responses in direct marketing models. In the context of response modeling, Bodapati and Gupta (2004) demonstrate that the discretization decreases bias and increases accuracy. In contrast to forecasting the annual number of product purchases individually, the managerial challenge is to predict whether this number exceeds a pre-defined threshold to determine the value of a discretization function d(y), which they define as follows:

$$d(y) = \begin{cases} 0 & \text{if } y \in (0, y_{threshold}] \\ 1 & \text{if } y \in (y_{threshold}, \infty) \end{cases} \qquad (7)$$

with $y_{threshold}$ as the pre-defined value of the absolute number of purchases as a typical example of a marketing application. From here, supervised classifiers can be built on the resulting function. Note that the discretization can be based on a revenue variable as well.

We extend equation (7) for revenue uplift modeling. To that end, we propose to first obtain $z_{i,rw}$ through applying CRVTW and to then convert $z_{i,rw}$ into a dichotomous response variable $z_{i,rg}$ through equation (8). Hence, rather than predicting $z_{i,rw}$ with regression models, a second response modification creates a binary response variable to be modeled using SML algorithms for classification. Thus, RDT extends the CRVTW transformation with the discretization scheme

$$z_{i,rg} = \begin{cases} 0 & \text{if } z_{i,r} \in (-\infty, 0] \\ 1 & \text{if } z_{i,r} \in (0, \infty) \end{cases} \qquad (8)$$

with $z_{i,rg} \in \{0,1\}$. A crucial difference between (8) and the discretization of Bodapati and Gupta (2004) is that the response variable in (8) has been pre-transformed and that negative numbers are captured in $z_{i,rg}$ as control group purchasers are included. This points out that in $z_{i,rg}$ information related to the treatment and control group is provided which is imperative to adopt an uplift modeling approach. In summary, uplift under RDT can be estimated by:

$$Uplift_{i,r}^{RDT} = E(z_{i,rg}|X_i). \qquad (9)$$

The reason why we set the threshold to be zero in equation (8) is related to the analytical objective in the context of uplift modeling. The definition of a "failure" is $z_{i,rg} = 0$ and includes non-incentivized purchasers ($z_{i,r} = -Y_{i,r}$), incentivized non-purchasers ($z_{i,r} = 0$), and non-incentivized non-purchasers ($z_{i,r} = 0$). In contrast, the definition of "success" relates to customers who have purchased a product with the causal connection to the campaign ($z_{i,r} = +Y_{i,r}$).

### 3.2.2 Covariates Transformation

Table 1 indicates alternative strategies that entail transformations of the data input space. Although not the methodological focus of this paper, corresponding covariates transformation approaches also provide a framework to perform revenue uplift modeling using SML algorithms. The interaction term method (ITM) by Lo (2002) and the treatment-covariates interactions approach (TCIA) by Tian et al. (2014) both prepare dichotomous response variables. However, extensions to continuous responses for accommodating revenues are straightforward. Due to the popularity of ITM in the uplift literature, we add this covariates transformation to the empirical study of the paper and briefly review its methodology.

The ITM approach uses the binary treatment indicator variable $T_i$ and introduces an interaction term $X_i \cdot T_i$ into the prediction model. Then, revenue uplift can be predicted as:

$$Uplift_{i,r}^{ITM} = f(X_i, T_i, X_i \cdot T_i). \tag{10}$$

Lo (2002) builds two equivalent uplift models on the treatment and control samples, whereby the covariate space in each sample was extended with the interaction term. To calculate uplift, the scores of the two models are subtracted. While Lo (2002) uses logistic regressions, his approach can accommodate arbitrary SML algorithms. Reviewing (10), it is apparent that the additional interaction term modifies the predictions of treated observations in contrast to control observations.

### 3.3 Two-Stage Revenue Uplift Models

We introduce novel models to account for zero-inflated revenue distributions. The simplest form of a zero-inflated model is a two-stage model where a classification model is built to predict whether the response will be zero or not (e.g., to identify purchase incidents) and then a regression model is built on non-zero responses to predict the actual response value (e.g., a buyer's purchase volume). More formally, a classification model estimates $E(Y_{i,c}) = f_c(X_i)$ and a regression model predicts $E(Y_{i,r}[Y_{i,r} > 0]) = f_r(X_i)$.

We multiply the separate model forecasts to calculate an expected purchase amount. For hurdle models (Mullahy, 1986), decomposing the log-likelihood into two independent parts and maximizing them separately is a valid approach as the estimated probabilities of both parts can be formulated as an additive term (see (Smithson & Merkle, 2013) (Hofstetter et al., 2016)). The same holds for zero-inflated Poisson regressions (Lambert, 1992). As we use an even simpler linear regression model for the second stages of our two-stage models, we do not see a structural difference to hurdle models and confirm the validity of separate model estimation for our two-stage models. The occurrence of two separate, sequential customer decisions in our e-commerce context further reinforces this argumentation.

The methodological setup of two-stage models differs across revenue uplift strategies due to varying response types. More specifically, the transformation of CRVTW prepares a response variable with positive (negative) values for buyers from treatment (control) groups and zero values for non-purchasers, see equation (5). To this end, we apply a classification model to identify purchasers (i.e., customers with a revenue value larger than zero) and a regression model on the transformed revenue response. RDT outputs a binary response variable, see equation (8), and can thus be can be considered a special case to two-stage models. To this end, we fit a classification model on the transformed response and a regression model on the revenue variable to predict purchase volumes. In terms of RDT, we further introduce the synthetic minority over-sampling technique (SMOTE), which is a popular approach to handle imbalanced class distributions by over-sampling the minority response class (i.e., creating artificial observations) and under-sampling the majority response class (Chawla et al., 2002). Hence, we build a SMOTE model on the discretized revenue response and use a classification model for the prediction task.

For the indirect revenue uplift strategy, following equation (4), we consider a two-stage model on the treatment sample and a separate two-stage model on the control sample. For each of these models, we use a classification model to forecast purchase incidents and a regression model to predict purchase volumes and multiply the respective predictions to derive sample-specific scores. We then subtract the score of the control group models from the score of the treatment group models. Finally, ITM-based two-stage models consider an additional interaction term for the second stage regression models on the treatment and control samples, see equation (10).

## 4 Campaign Profit for Uplift

Subsequently, we develop an easily interpretable profit decomposition to measure the incremental profit impact of a causal model. The novel measure details costs related to a couponing setting but can be applied in other marketing settings as well. Hansotia and Rukstales (2002b) develop a break-even decision rule that is estimated by expected profit subtracted by costs per contact. In contrast to their rule, we decompose campaign profit for uplift into expected costs associated with the marketing campaign and differential gain in revenue, define profit as a function of lift, distinguish different types of costs, and facilitate flexibility in the choice of arbitrary direct and indirect uplift models. To derive this profitability, we start from the marketing campaign profit measure, $\Omega$, of (e.g., Lessmann et al., 2019)

$$\Omega^{Lift} = N * \tau * (\pi_+ * l(\tau) * \delta - \varepsilon), \tag{11}$$

with N referring to the total population of customers, $\tau$ the share of targeted customers, $\pi_+$ the fraction of customers who respond to the campaign, $\gamma(\tau)$ the lift index which can be expressed as $\gamma(\tau) = \pi_\tau/\pi_+$ where $\pi_\tau$ indicates the share of actual positive reactions within the campaign target group (e.g., Neslin et al., 2006), and $\delta$ and $\varepsilon$ as revenue and costs, respectively. Hence, campaign profit can be re-formalized as $\Omega^{Lift} = N * \tau * (\pi_+ * \pi_\tau / \pi_+ * \delta - \varepsilon) = N * \tau * (\pi_\tau * \delta - \varepsilon) = N_\tau * \pi_\tau * \delta - N_\tau * \varepsilon$ with $N_\tau = N * \tau$ for simplification.

Martens and Provost (2011) express campaign profit as a function of the lift measure. To generalize their measure to uplift modeling, we develop a profit equation that captures incrementality. We therefore modify the term $N_\tau * \pi_\tau * \delta$ by $N_\tau * \pi_\tau * \delta_\tau - N_\varsigma * \pi_\varsigma * \delta_\varsigma$, with $N_\varsigma, \pi_\varsigma$ as control group equivalents to the quantities $N_\tau, \pi_\tau$ and $\delta_\tau$ and $\delta_\varsigma$ as the average revenue in the treatment and control groups, respectively. With this, we add necessary control group information to the revenue side of the equation. Hence, we express campaign profit for uplift as $\Omega^{Uplift} = (N_\tau * \pi_\tau * \delta_\tau - N_\varsigma * \pi_\varsigma * \delta_\varsigma) - N_\tau * \varepsilon$.

The term $N_\tau * \varepsilon$ on the cost side implies that only contact costs occur. However, specific campaigns may require a more comprehensive cost model. We, therefore, generalize uplift campaign profit to

$$\Omega^{Uplift} = (N_\tau * \pi_\tau * \delta_\tau - N_\varsigma * \pi_\varsigma * \delta_\varsigma) - \varepsilon_{treatment} \tag{12}$$

with $\varepsilon_{treatment}$ reflecting all treatment-related costs of a targeting campaign. First, $\varepsilon_{treatment}$ embodies contact costs, $\varepsilon_{contact}$. They occur whenever a customer receives a treatment and thus depend on the number of customers targeted. Contact costs are thus a function of the fraction targeted and $c_{unit}$ stating the constant unit costs for performing a promotional action. This suggests $\varepsilon_{contact}(\tau) = N_\tau * \varepsilon_{unit}$. Depending on the area of application, unit costs vary from zero or near-zero (e.g., automated e-couponing) to several euros per transaction (e.g., outbound call campaigns with personal customer support).

In addition to contact costs, a second component of $\varepsilon_{treatment}$ in equation (12) refers to the costs of the incentive, which the campaign offers to solicited customers. Such costs are especially relevant for applications where a financial discount plays a critical role in persuading customers to buy (e.g., couponing) and occur as soon as a treated customer accepts the marketing offer. In the following, we focus on campaigns that offer a relative discount and define $\rho \in \{0.05, \dots, 0.95\} \subseteq \mathbb{R}$ to be the corresponding price reduction. Typically, promotion values are in intervals of five (i.e., 5%, 10%, etc.). The choice of a relative discount has been suggested by the marketing agency that has provided the data for the empirical study. It does not constrain the campaign profit formulation and can be exchanged with an absolute discount value or other cost factors when needed. Discounts limit revenue to the exact extent of the financial value of the incentive that has been used for the promotion. Hence, we describe incentive costs as $\varepsilon_{incentive}(\tau, \delta_{\pi_+}, \rho) = \rho * \sum_{i=1}^{N_\tau} \delta_{\pi_+,i}$ In our example, the coupon provides a benefit of saving ten percent off the targeted customer's basket values, i.e., $\rho = 0.1$. In contrast to contact costs, incentive costs depend on the individual customer's shopping basket size.

Based on contact and incentive costs, we modify campaign profit as stated in equation (12) as follows:

$$\Omega^{Uplift} = (N_\tau * \pi_\tau * \delta_\tau - N_\varsigma * \pi_\varsigma * \delta_\varsigma) - \varepsilon_{treatment}$$
$$= (N_\tau * \pi_\tau * \delta_\tau - N_\varsigma * \pi_\varsigma * \delta_\varsigma) - (\varepsilon_{contact}(\tau) + \varepsilon_{incentive}(\tau, \delta_{\pi_+}, \rho))$$
$$= (N_\tau * \pi_\tau * \delta_\tau - N_\varsigma * \pi_\varsigma * \delta_\varsigma) - N_\tau * \varepsilon_{unit} - \rho * \sum_{i=1}^{N_\tau} \delta_{\pi_+,i} \quad (13)$$

Equation (13) presents the fundament for marketing investment decisions on a decile-level. We assess revenue uplift models in terms of equation (13) and several other performance indicators and compare their financial impacts to several benchmarks.

## 5 Experimental Setup

### 5.1 Campaign and Data

An online marketing agency supports this research by providing data to test the effectiveness of revenue uplift modeling. The marketing campaign is based on a real-time targeting process. The purpose of uplift models is to score visitors of an online shop according to their browsing behavior. Targeted customers are offered a ten percent discount off their shopping basket. This setting displays typical characteristics of targeted marketing in that the coupon should only be offered to undecided and persuadable customers to avoid a waste of marketing resources.

The marketing agency uses the following process to gather data. A propensity model pre-screens online shop visitors. This model consists of a random forest classifier and predicts purchase expectations (Baumann et al., 2019). Visitors with a high likelihood to buy are not eligible for a discount. The remaining visitors are randomly distributed into a treatment and control group at a ratio of 3:1. Treatment group visitors are offered the discount, which they can use in the current browsing session. Purchases and basket values of treatment and control group visitors are recorded and available in our data. Although the filtering mechanism, which the marketing agency imposes to avoid soliciting likely buyers, only excludes a very small fraction of visitors from the chance of receiving a coupon, it creates a quasi-experimental setting (e.g., Armstrong & Patnaik, 2009). While a truly randomized trial was preferable to estimate the causal effect of the digital coupon, the focus of this paper is to examine the relative effectiveness of alternative uplift modeling strategies. The data facilitates a comparison of different models on equal ground, which suggests that it is suitable for the study.

The data set comprises 2,951,313 observations and 60 variables and facilitates a large-scale empirical analysis. Since the data comes from twenty-five European online shops, it also provides a broad view across different product assortments and visitor patterns. We further detail the employed e-commerce data in the online appendix (see Figure A1.1). Each observation represents an individual session; the whole customer shopping journey from shop access until leave. The data consists of variables that relate to the time, views, baskets, prior visits and technical characteristics of customers. Furthermore, meta-variables are included.

While the data comprises mainly discrete and continuous numerical variables, some variables are categorical. The online appendix provides an overview of the variables and detailed descriptions (see Table A1.1).

Uplift statistics such as the signal-to-noise ratio are commonly calculated to clarify campaign effectiveness (Kane et al., 2014). Such measures relate to conversion uplift and compare responders from the treatment/control groups to determine the initial (conversion) uplift of the data. Apart from this, we further add analogous information on aggregated customer spend to capture initial revenue uplift. Therefore, Table 2 provides corresponding information in terms of (i) the number of customers in the treatment and control groups, (ii) the number of customers having purchased at least one product as well as the incurred spend volume in total for each of the groups and in average per customer, and (iii) the uplift based on the group differences in conversion rates and summed magnitudes of purchases.

**Table 2: Descriptive uplift statistics of experimental data**

| Group Affiliation | Share [%] | Customer Sessions | Purchasers | Conversion Rate [%] | Conversion Uplift [%] | Revenue [€] | Revenue Per Person [€] | Revenue Uplift [€] |
|---|---|---|---|---|---|---|---|---|
| **Treatment** | 74.9 | 2,210,190 | 162,570 | 7.35 | 0.16 | 178,216 | 0.08 | 0.05 |
| **Control** | 25.1 | 741,123 | 53,340 | 7.19 | | 58,149 | 0.03 | |
| *Total* | *100* | *2,951,313* | *215,910* | *-* | *-* | *236,365* | *-* | *-* |

Conversion uplift has also been expressed as the average treatment effect in previous studies (e.g., (Guelman et al., 2015a)) and is not statistically significant for our data (p-value of 5.93e-06) as determined by Pearson's chi-squared tests. Therefore, with a conversion and (normalized) revenue uplift of 0.16% and 0.05€, respectively, the overall uplift signal is rather low.

## 5.2 Uplift Models and Learning Algorithms

Table 3 summarizes the modeling strategies to develop revenue uplift models, their category and formalization, as well as the underlying learning algorithm we consider in this paper. We sort uplift strategies according to their formalization (i.e., equation number). Apart from own preliminary research (Gubela et al., 2017), to the best of our knowledge, uplift literature has not yet studied revenue uplift strategies. The interaction term method and the indirect revenue uplift strategy emerge as straightforward extensions of their conversion uplift modeling equivalents (e.g., Devriendt et al., 2018; Kane et al., 2014). The focus of this paper is thus on the two response transformations for uplift modeling, which we highlight in bold font in Table 3.

**Table 3: Revenue uplift strategies**

| Revenue Uplift Strategy | Category | Model Formalization | Learning Algorithm |
|---|---|---|---|
| **Continuous Response Variable Transformation with Weightings (CRVTW)** | Response Transformation | (7) and (8) | Regression |
| **Revenue Discretization Transformation (RDT)** | Response Transformation | (10) and (11) | Classification |
| Interaction Term Method (ITM) | Covariates Transformation | (12) | Regression |
| Indirect Revenue Uplift Strategy (INDIRECT) | Two Model Approach | (4) | Regression |

Implementing a conversion or revenue uplift strategy requires a classification or regression learning algorithm. As RDT produces a binary response variable, see equation (8), implementing an RDT revenue uplift model requires an underlying classification algorithm. In contrast to this, regression algorithms implement all revenue uplift strategies except RDT. For the classification task, we consider tree-based algorithms (e.g., random forest, gradient boosting, and extremely randomized trees), k-nearest neighbors, support vector machines, and several others. For the regression task, we also use random forest and support vector machines, but further add neural networks and some more specific algorithms such as Lasso LARS (LL) and Theil-Sen regressors (TSR). Table A1.2 and Table A1.3 from the online appendix provide details on the algorithms' meta-parameter specifications, candidate settings and the number of related models for the classification and regression tasks, respectively. These numbers emerge from estimating a candidate model for every possible combination of meta-parameter settings per learning algorithm during model selection (e.g., Lessmann et al., 2015). Hastie et al. (2009) provide a detailed description of the classification and regression methods.

Apart from semi- and non-parametric models, we further implement the revenue uplift strategies employing parametric models. The one-stage revenue uplift models are logistic regression (LogR) models, linear discriminant analysis (LDA) classifiers without shrinkage and ordinary least squares (OLS) regression models. Hence, we apply LogR and LDA with the RDT strategy to solve a classification task and OLS with the remaining strategies to solve a regression task and calculate uplift scores. We choose these algorithms due to their parametric properties and their popularity in the machine learning community. We further use the same learning algorithms for the two-stage models. Therefore, for all strategies, we consider LogR and LDA as classification models for the first stage and OLS as a regression model for the second stage. For RDT, we further consider SMOTE. We summarize all models in Table 4.

**Table 4: Parametric revenue uplift models with one- and two-stage base learners**

|  | RDT | CRVTW | ITM | INDIRECT |
|---|---|---|---|---|
| **One-Stage** | LogR | OLS | OLS | OLS |
|  | LDA | | | |
| **Two-Stage** | LogR, OLS<br>LogR (SMOTE) | LogR, OLS | LogR, OLS | LogR, OLS |
|  | LDA, OLS<br>LDA (SMOTE) | LDA, OLS | LDA, OLS | LDA, OLS |

## 5.3 Parameter Tuning and Performance Evaluation

We randomly partition the data into a training set (~40% or 1,220,805 observations), a validation set (~30% or 915,606 observations), and a hold-out test set (~30% or 915,579 observations). Each of these sets includes browser sessions from the treatment and control group. We use the training set to develop uplift models using the learning algorithms of Table A1.2 and Table A1.3 (see online appendix). To illustrate this, consider, for example, the first learner from Table A1.3, which is a LL model. To implement an indirect revenue uplift strategy according to (4), we estimate one LL model from the treatment group observations of the training set and another regression model from the control group observations of the training set. We then score observations in the validation set with both regression models and calculate the differences between the two models' scores. This difference represents an uplift score, which we use for assessment. We repeat the process for the other revenue uplift strategies (Table A1.3) and develop corresponding uplift models based on LL regression. Next, we consider a different learning algorithm and repeat; considering different meta-parameter settings whenever a learner exhibits meta-parameters.

The validation set facilitates identifying the best combination of a learning algorithm and meta-parameter configuration per revenue uplift strategy. For example, we may find the best ITM revenue uplift model to come from a ridge regression (Ridge) learner with $\alpha = 0.01$ (see Table A1.3). We use this specification and re-estimate the best ITM revenue uplift model from the union of the training and validation set (covering about 70% of the whole data) to obtain our final ITM revenue uplift model, which enters subsequent comparisons to other uplift strategies and benchmarks on the test set. To ensure the robustness of results, we repeat the random partitioning of the data into training, validation, and test set – and thus all intermediate steps described above for one random partitioning – ten times. As we do not perform hyperparameter optimization for the analysis of one-stage vs. two-stage models, we thereby perform model training on both the training and validation sets and predict on the hold-out test set.

Due to the fundamental problem of causal inference (Holland, 1986), which points to the impossibility of simultaneously targeting and not targeting the same customer, our analyses require uplift-specific performance measures. The Qini coefficient and Qini curves, generalizations of the Gini coefficient and gain charts for uplift models, have been developed for this purpose (Radcliffe, 2007). Our focus on revenue uplift modeling implies an extension of such measures to a revenue setting. To this end, the Revenue Qini coefficient $Q_r$ considers the incremental gain in revenue, derived by calculating the decile-wise difference in the revenue metric between treatment and control group observations. $Q_r$ is defined as the area between the respective Revenue Qini curve, which illustrates the performance of a revenue uplift model for each of the ten targeted deciles, and a diagonal line representing random targeting (Radcliffe & Surry, 2011).

To emphasize the flexibility in the choice of financial performance indicators for practitioners, we use varying expressions of $Q_r$ and Revenue Qini curves for different parts of our empirical study. More specifically, these comprise revenue per person and total revenue, normalized revenue for comparisons across experimental settings, and a scaled version of $Q_r$ to dissolve the dependency on the sample size (Radcliffe & Surry, 2011). Note that from a perspective of corporate practice, Revenue Qini curves better prepare operational decision-making compared to $Q_r$ as they summarize model performance on a more granular level (i.e., per targeted decile). Therefore, we add different measures of Revenue Qini curves to further increase the quality of decision-making. For more detailed descriptions of Qini-related performance measures, we refer to the online appendix. Apart from $Q_r$ and Revenue Qini curves, we assess campaign effectiveness in terms of campaign profit for uplift as derived in Section 4 and model computation times.

# 6 Empirical Analysis of Uplift Transformations for Revenue Uplift Models

We split the evaluation of the proposed revenue uplift transformations into two parts. Revenue uplift transformations require an underlying learning algorithm. The first part clarifies, for each individual transformation, whether this algorithm should be a standard SML algorithm or a two-stage model. We restrict this part to linear models. Drawing upon corresponding results, part two assesses several configurations of our revenue uplift transformations using a rich set of SML algorithms, either in a one- or a two-stage regime, and together with hyperparameter tuning. The second analysis identifies the best way to implement revenue uplift modeling for the available data. Finally, we empirically substantiate our claim that revenue uplift models give higher campaign revenue than uplift models for conversion. For better readability, the online appendix provides a list of all acronyms used in the empirical analysis.

## 6.1 Two-Stage Models for Zero-Inflated Revenue Distributions

We report empirical results obtained from instantiating revenue uplift transformations using one- or two-stage models in Table 5. We assess different models in terms of the Revenue Qini coefficient $Q_r$, which we measure on the test set. Higher Qini values indicate better performance. In Table 5, we concentrate on linear models and use linear and logistic regression to develop uplift models. This way, we circumvent the need to tune algorithmic hyperparameters which simplifies the analysis. The one-stage approach can then be considered the standard way to develop an uplift model, whereas the two-step approach addresses the skewed distribution of the revenue target variable. We highlight the preferable model for each strategy using a bold font.

**Table 5: Revenue Qini coefficient of parametric revenue uplift models with one- and two-stage base learners**

| | | Revenue Qini coefficient | |
|---|---|---|---|
| Strategy | Model | One-Stage | Two-Stage |
| **RDT** | LogR | **0.021** | |
| | LogR (SMOTE) | | 0.011 |
| | Two-Stage (LogR, OLS) | | 0.020 |
| | LDA | 0.017 | |
| | LDA (SMOTE) | | 0.007 |
| | Two-Stage (LDA, OLS) | | 0.010 |
| **CRVTW** | OLS | | **0.159** |
| | Two-Stage (LogR, OLS) | | 0.149 |
| | Two-Stage (LDA, OLS) | | 0.144 |
| **ITM** | OLS | | **0.136** |
| | Two-Stage (LogR, OLS) | | 0.118 |
| | Two-Stage (LDA, OLS) | | 0.049 |
| **INDIRECT** | OLS | | 0.146 |
| | Two-Stage (LogR, OLS) | | 0.178 |
| | Two-Stage (LDA, OLS) | | **0.187** |

Table 5 suggests puts the two-stage approach into perspective. For most of the modeling strategies, it does not improve results over the classic one-stage approach. For RDT, simple one-stage implementations using LogR and LDA perform much better than their SMOTE and two-stage model counterparts. This is also true for ITM and CRVTW, where the OLS implementation outperforms two-stage alternatives, independent on whether the latter is implemented using LogR or LDA. Only the indirect revenue uplift strategy's benefits from addressing skewed revenue distributions using a two-stage model, whereby the LDA-based implementation outperforms the LogR-based one. The online appendix offers a more detailed view on the performance of alternative models in terms of cumulative incremental normalized revenue per person along targeted deciles using Revenue Qini curves (see Figure A2.1). Corresponding results agree with Table 5 to a large extent. Based on these results, we develop RDT, CRVTW, and ITM uplift models using standard SML algorithms in subsequent experiments. For the indirect uplift model, we pursue a two-stage approach.

## 6.2 Analysis of Revenue Uplift Strategies and Learning Algorithms

This part of the analysis aims at evaluating the performance of the proposed revenue uplift strategies RDT and CRVTW as well as that of revenue uplift models based on ITM and the indirect approach. The comparison of alternative approaches also considers the interaction between a revenue uplift strategy and the underlying SML algorithm. We measure the performance of each revenue uplift strategy across learning algorithms in terms of $Q_r$ which we express as the revenue per customer from the test set observations. Table 6 and Table 7 depict the empirical results for classification and regression algorithms, respectively. More specifically, we report the results of the best model per revenue uplift strategy and learning algorithm after hyperparameter tuning on the validation set. All scores are mean values over ten cross-validation iterations. Recall that RDT is the only revenue uplift strategy that advocates the use of classification algorithms, whereas other revenue uplift transformations use regression algorithms to estimate a continuous target variable. In both tables, we highlight the best and second-best models in bold and italic font, respectively, and the third-best model with an underscore. We further add a revenue response model to Table 7.

**Table 6: Revenue Qini coefficient of classification-based RDT uplift model across learning algorithms**

| Revenue Strategy | Classification Algorithm | | | | | | | |
|---|---|---|---|---|---|---|---|---|
| | ERT | GBC | KNN | LDA WS | LogR | QDA | RFC | SVC |
| **RDT** | **0.745** | 0.555 | 0.462 | 0.376 | 0.461 | 0.266 | *0.634* | 0.242 |

**Table 7: Revenue Qini coefficient of regression-based revenue uplift models across learning algorithms**

| Revenue Strategy | Regression Algorithm | | | | | |
|---|---|---|---|---|---|---|
| | LL | MLP | RFR | Ridge | SVR | TSR |
| **CRVTW** | -0.136 | 0.053 | 0.282 | 0.218 | 0.156 | 0.257 |
| **ITM** | -0.049 | 0.041 | -0.139 | **0.572** | -0.114 | 0.283 |
| **Indirect Approach*** | 0.351 | 0.026 | 0.038 | 0.257 | 0.108 | 0.024 |
| **Response Modeling** | -0.138 | 0.071 | <u>0.389</u> | *0.473* | 0.225 | 0.304 |

* According to Table 5, we implement the indirect approach using a two-stage model. We use LDA for the first stage and consider several regression algorithms for the second stage, which undergo hyperparameter tuning.

RDT gives the best results together with an extremely randomized tree base model. The top-three classifiers for RDT ground on a tree-learning, which supports the popularity of decision trees in previous work (see Table 1). A tree-based learner, random forest regression, provides the best result for CRVTW, whereas ITM and the indirect approach work best when implemented using ridge regression. Interestingly, a revenue response model outperforms several uplift strategies in terms of $Q_r$ although it disregards the causal effect of the treatment on customer behavior. We attribute this result to the data with relatively small revenue uplift (see Table 2). However, based on the obtained results, we emphasize that a careful selection of a revenue uplift strategy is crucial. While the response model outperforms several uplift model configurations, the best uplift model, ITM based on ridge regression, outperforms the best response model with a twenty-one percent margin.

Comparing the performance of the best classifiers with the best regressors, we observe the highest performance for the RDT-based extremely randomized tree, which is 30.2% better than the ITM-based ridge regression. RDT also contributes the second-best and fourth-best result across Table 6 and Table 7 using, respectively, a random forest and gradient boosting implementation. CRVTW, on the other hand, does not achieve any of the top three ranks. This is remarkable since RDT is based on a discretization of the continuous target variable that emerges from a CRVTW uplift transformation. Substantially better performance of RDT supports results of Bodapati and Gupta (2004) and further extends these to uplift settings.

Table 6 and Table 7 offer an aggregated view on model performance. Interested readers find more detailed results in the online appendix, where we report the distribution of model performance across cross-validation iterations (see Figure A2.2). Corresponding results indicate that RDT provides higher uplift and less variation compared to other uplift strategies. Overall, we find tree-based RDT algorithms to outperform their competitors. We further secure this conclusion with a decile-wise analysis of alternative revenue uplift strategies using Revenue Qini curves. More specifically, we consider the revenue uplift strategies with their best underlying base learners as per Table 6 and Table 7. Corresponding results are available in the online appendix (see Figure A2.3) and confirm the RDT model to outperform other models on most deciles. We also observe CRVTW to perform inferior to other strategies such as a simple revenue response model.

To summarize previous results and provide a holistic picture of model performance, we perform a multi-criteria evaluation. This evaluation includes the Qini curve values of important lower deciles due to fixed budget constraints, the Qini curve value as an average across deciles and a weighted Qini curve value based on the procedure proposed by Ling and Li (1998), which weights decile-wise results according to their importance for marketing. More formally, $Q_{wr} = (0.9 * Q_{1,r} + 0.8 * Q_{2,r} + \cdots + 0.1 * Q_{9,r})/\sum_i Q_{i,r}$ with $wr$ indicating weighted incremental revenue and $i = 0, 1, \ldots, 9$ representing an index of the respective targeting deciles. Contrary to $Q_r$, which takes random targeting into account but disregards decile-specific model performance, the Qini curve-related measures do not consider random targeting but focus on different aspects of incremental, cumulative model performance (e.g., model performance in terms of the first and third deciles, respectively). Therefore, they are more suitable to guide operational decision-making. Table 8 reports the results of the multi-criteria evaluation. For each dimension, the best, second-, and third-best model is highlighted in boldface, italic font, and with an underscore, respectively. We also report the average rank of an uplift modeling strategy across performance measures.

**Table 8: Multi-criteria evaluation of revenue models**

| Revenue Models | Top 10% Qini Curve | Top 30% Qini Curve | Averaged Qini Curve | Weighted Qini Curve | Average Rank |
|---|---|---|---|---|---|
| **RDT (ERT)** | **1,470** | **1,505** | **1,599** | **6,806** | 1 |
| CRVTW (RFR) | 468 | 892 | 1,119 | 4,226 | 4 |
| ITM (Ridge) | <u>1,034</u> | *1,431* | *1,424* | *6,313* | 2 |
| INDIRECT_TS (Ridge) | 310 | 804 | 1,094 | 3,928 | 4 |
| RESPONSE (Ridge) | *1,191* | <u>1,191</u> | <u>1,323</u> | <u>5,563</u> | 3 |

Table 8 provides strong evidence in favor of the proposed RDT revenue uplift model. It performs best on all dimensions and achieves the overall first rank. Due to being ranked second in both the top 30% Qini curve, the averaged Qini curve and the weighted Qini curve, ITM is the second-best strategy, followed by the response model-based ridge regression with three third-best ranks and a second-best rank on the top 10% Qini curve. Given substantial differences between the simple revenue response model and RDT as well as ITM in terms of three out of four assessment criteria, Table 8 makes a strong case for revenue uplift modeling and the cruciality of a causal approach to target marketing activities. Empirically, revenue response modeling does not prove to be a viable alternative. CRVTW and the indirect approach do not achieve a good rank on any performance dimension and share the last position in the multi-criteria evaluation.

### 6.3 Revenue Versus Conversion Uplift Models

We complete the analysis of revenue uplift transformation with a comparison of selected strategies to classical conversion uplift models. In line with Table 8, we chose RDT and ITM, as the best revenue uplift strategy based on classification and regression models, respectively for the comparison. For conversion uplift models, we draw on previous results of Kane et al. (2014) that identify ITM and LWUM as competitive modeling strategies for conversion uplift. These approaches require an underlying base learning algorithm for classification. We consider the same algorithms as in Table A1.2 from the online appendix to implement conversion uplift models. Comparative results are available in Table 9. As in the revenue uplift case, we report results only for the best conversion model, which we determine through comparing alternative classification algorithms and hyperparameters on the validation data. For completeness, the table also reports results for an ordinary response model. We mark the most effective model for each decile in italic font and highlight the overall top three strategies in bold.

**Table 9: Comparison of revenue and conversion uplift modeling**

| Revenue Strategy | Model Focus | Model | Normalized Revenue Per Decile | | | | | | | | | |
|---|---|---|---|---|---|---|---|---|---|---|---|---|
| | | | Dec. 1 | Dec. 2 | Dec. 3 | Dec. 4 | Dec. 5 | Dec. 6 | Dec. 7 | Dec. 8 | Dec. 9 | Dec. 10 |
| **RDT** | Revenue | ERT | *1,469.8* | 1,199.5 | *1,504.8* | 1,637.5 | 1,396.2 | 1,668.2 | ***1,906.6*** | ***1,891.9*** | 1,716.6 | 1,682.8 |
| **ITM** | Revenue | Ridge | 1,034.2 | *1,618.0* | 1,430.7 | *1,637.6* | 1,468.0 | 1,279.8 | 1,397.2 | 1,436.4 | 1,512.4 | 1,682.8 |
| | Conversion | SVC | 271.8 | 993.5 | 1,207.7 | 1,181.3 | 1,398.7 | 1,532.6 | 1,717.0 | 1,603.1 | 1,670.2 | 1,682.8 |
| | | KNN | 412.4 | 747.8 | 913.4 | 985.7 | 1,212.4 | 1,489.5 | 1,488.6 | 1,842.6 | ***1,839.5*** | 1,682.8 |
| **LWUM** | Conversion | RFC | 1,369.3 | 1,278.3 | 1,332.2 | 1,386.1 | *1,643.1* | *1,792.9* | 1,867.9 | 1,829.2 | 1,817.3 | 1,682.8 |
| | | KNN | 452.8 | 828.6 | 952.1 | 1,288.2 | 1,396.7 | 1,534.4 | 1,613.0 | 1,788.0 | 1,839.4 | 1,682.8 |
| **Response Modeling** | Conversion | LogR | 1,339.8 | 1,224.3 | 823.4 | 1,372.4 | 1,272.7 | 1,330.3 | 1,350.5 | 1,384.6 | 1,509.0 | 1,682.8 |
| | | RFC | 1,456.2 | 965.2 | 1,353.7 | 1,330.4 | 1,393.0 | 1,575.8 | 1,725.9 | 1,723.6 | 1,665.3 | 1,682.8 |

The first finding of Table 9 is that the two best results with the highest incremental revenue come from RDT. This confirms previous findings and the suitability of the RDT modeling strategy. For the data under study, we observe large campaigns targeting 70% of the customer base to give the highest normalized revenues. We attribute this pattern to the specific data used for the comparison. In general, marketers prefer targeting a small fraction of customers. Therefore, it is appealing to observe revenue uplift strategies to also provide the best results for each of the first four deciles. In this regard, Overall, Table 9 confirms the suitability of revenue uplift modeling. Corresponding modeling strategies such as RDT increase campaign revenue to a larger extent than traditional uplift models for conversion. From a marketing perspective, this finding might not come as a surprise, as it stresses the fact that customers differ in their spending. A targeting model should take this heterogeneity into account. However, we reiterate that much prior literature on machine learning-based targeting models does not consider uplift models at all and that the few uplift modeling studies predominantly use conversion modeling.

# 7 Analysis of RDT Uplift Strategy Against Causal Machine Learning

Previous analysis has identified RDT as the most suitable revenue uplift strategy for the available e-commerce data. We have also observed this strategy to perform best when implemented with an ERT base learning algorithm and obtained a set of optimal hyperparameters for the base learner. To set the performance of this specific configuration, our best revenue uplift model, into context, we compare it to several recently proposed causal machine learning algorithms. The causal learning algorithms include causal trees and their successor causal forests (Athey & Imbens, 2016; Athey et al., 2019), causal boosting (Powers et al., 2018), causal BART (Hill & Su, 2013; Hill, 2011), and an x-learner (Künzel et al., 2019), which we implement using a random regression forest as base learner. These benchmarks facilitate estimating customer-level CATE for continuous target variables and thus revenue uplift modeling.

Software to apply the causal machine learning algorithms are publicly available. We detail the employed packages, our hardware and how we specified the causal machine learning algorithms in the online appendix (see Subsection A3.1). Each of the benchmarks has shown strong performance in previous evaluations (see, e.g., Knaus et al., 2018 as well as the above studies, which have introduced the algorithms). However, we are not aware of a previous evaluation in an uplift modeling context. Marketing data sets are typically high-dimensional and comprise a large number of observations. To shed light on the scalability of available software libraries for causal machine learning algorithms, subsequent analysis reports computation time alongside predictive performance. In addition to Qini scores and runtime measurements, we use our proposed incremental profit measure, as derived in equation (13), to assess RDT and causal machine learning benchmarks. Corresponding results clarify the degree to which revenue uplift models add to the bottom line and whether differences in the predictive performance of alternative models are managerially meaningful. Given that the specific configuration of RDT emerges from Section 6, using the same data for the following comparison would give RDT an unfair advantage. Therefore, we use a fresh set of data for the comparison. The new data comes from three online shops that were not part of the original, cross-shop data set. These shops are new clients from our data provider. Thus, the additional data was also captured in a different period and includes an extended set of covariates. Descriptive statistics and some more detailed information on the new data are available in the online appendix (see Table A3.1). Subsequently, we refer to the new data sets using the acronyms BAT, FA, and FB. We apply the RDT uplift model as specified in Section 6 without further tuning to the new data sets. Considering structural and temporal differences between the cross-shop and the new data, the following evaluation may be considered an out-of-universe test of RDT.

Table 10 reports empirical results, which we obtain from a random subset of 100,000 observations from the BAT, FA and FB data sets. More specifically, we partition sampled observations into ten bins of 10,000 observations, stratify each bin into 70% training and 30% test data, and repeat the estimation of a model on the training data and assessment on the test data for each partition. In Table 10, we express $Q_r$ as total (non-normalized) revenue. Although the subsampling ensures the number of observations to be the same for each shop, we discourage from comparing Revenue Qini scores across data sets because the purchase volume and revenue uplift vary substantially across data sets as shown in the online appendix (see Table A3.1). Table 10 also reports computation times per data set. We use bold and italic font to highlight the best and second-best model per evaluation criteria and data set, respectively. All values represent averages across ten partitions.

**Table 10: Performance and computation times of causal machine learning algorithms**

| Causal machine learning algorithm | Revenue Qini coefficient | | | Computation times (sec) | | |
| --- | --- | --- | --- | --- | --- | --- |
| | BAT Data | FA data | FB data | BAT data | FA data | FB data |
| **RDT ERT** | **1.20** | **0.12** | **0.39** | 6.27 | *3.51* | 4.41 |
| **Causal Tree** | 0.15 | 0.02 | 0.21 | **0.59** | **0.54** | **0.51** |
| **Causal Forest** | 0.25 | 0.07 | *0.32* | *3.50* | 3.70 | *3.21* |
| **Causal Boosting** | 0.18 | 0.04 | 0.18 | 1,512.00 | 393.06 | 207.33 |
| **Causal BART** | 0.24 | *0.10* | 0.17 | 130.21 | 105.64 | 96.11 |
| **X-Learner RF** | *0.28* | 0.04 | 0.27 | 185.97 | 182.35 | 181.80 |
| Mean | 0.38 | 0.07 | 0.26 | 306.42 | 114.80 | 82.23 |

Before elaborating on our interpretation of Table 10, we note that a more detailed view on model performance is again available in the online appendix in the form of Revenue Qini curves (see Figure A3.1). Considering the aggregated (over deciles) Qini scores of Table 10, we find RDT to deliver the overall best revenue uplift model. It excels on the BAT data set where Qini scores are 4.29 times better than those of the second-best model (x-learner random forest). For example, the observed Revenue Qini coefficient of 1.20 implies that the summed revenue difference between treatment and control group customers across deciles is larger than the number of test set customers. RDT also performs consistently better than causal machine learning benchmarks on FA and FB. Performance gains are less substantial compared to BAT. However, improvements of 71.43% (FA) and 21.88% (FB) in terms of $Q_r$ over the strongest benchmark are still sizeable and confirm the promising performance of RDT. In appraising these results, it is important to note that we do not tune hyperparameters of the causal machine learning algorithms. Their application mimics how we use RDT using pre-specified hyperparameter settings. A tuning of hyperparameters is likely to alter empirical results for all techniques in the comparison. Therefore, we expect tuning to mainly change the level of results but not the relative order of competing algorithms.

Comparing the causal machine learners to each other, Table 10 confirms that causal forests outperform their predecessor, the causal tree, across all data sets. Overall, the causal tree appears less suitable for uplift modeling as it typically delivers the lowest Revenue Qini scores. However, it is important to note that the causal tree is by far the most efficient algorithm in the comparison. Estimating a model on a 10,000 observation sample requires only half a second on average. Overall, Table 10 indicates that causal forests might be the best causal machine learning algorithm for the focal data. Relative to other causal learners, its Qini scores are consistently high and the algorithm requires only a few seconds to estimate a model. The x-learner performs roughly as good as causal forests but is much slower. Contrary to recent literature that confirms the empirical effectiveness of causal boosting and causal BART (Dorie et al., 2019; Wendling et al., 2018), we find these methods less in favor of these methods. Especially causal boosting does not perform well in our comparison. Its Qini scores are consistently lower than those of the best causal machine learner and the algorithm needs by far the most time to estimate a model.

Knaus et al. (2018) also raise some concerns related to the scalability of some causal machine learning algorithms. We caution against over-emphasizing corresponding results of Table 10. We find it important to distinguish between the scalability of an algorithm and that of an implementation of an algorithm. The sharing of codes that emerge from research initiatives via public repositories is a great development. Experiments like that of Table 10 would not have been possible had the authors of the causal machine learners not released their codes to the public. One cannot expect such codes to be fully optimized and production-ready. For example, despite many empirical successes of the gradient boosting machine, it took years before highly efficient implementations of that algorithm became available. Our interpretation of observed results is thus that the runtimes of Table 10 indicate that the employed implementation of causal boosting is maybe less mature than that of, e.g., causal forests. Researchers and practitioners interested in experimenting with causal machine learning algorithms on large data sets might find this result useful to guide their choice of techniques. Finally, concerning computation times, we note that the proposed RDT approach displays competitive results. Although being less efficient than causal trees and causal forests, the observed runtimes do not raise concern against RDT. Again, its efficiency depends on the underlying learning algorithm. We choose ERT because it gave the best empirical performance in Section 6. However, RDT can easily accommodate other classifiers including high-performance implementations of, e.g., random forest or gradient boosting. Therefore, we are confident that scalability is not a problem for RDT.

To give a clearer view on the business impact of using alternative revenue uplift models, we use our profit decomposition to measure the financial return of e-couponing campaigns and subsequently report campaign profit for uplift among RDT and causal machine learning algorithms. Contact costs can be neglected for the data under study, which comes from e-couponing. Digital coupons can be issued at zero costs, for example through inserting a pop-up window in a browser session. However, we consider the costs associated with the discount value (i.e., incentive costs). We assume that treated buyers have activated the e-coupon as part of the checkout process. As before, profit is not normalized as we assess the effectiveness of the marketing campaigns on independent online shops. This allows us to assess the bottom-line impact of our proposed strategy for each of the three online shops. Still, the decile-wise values contain the attributes of incrementality and cumulativeness. Based on these preliminaries, we calculate campaign profit using equation (13) and report corresponding results in Table 11. To this end, Table 11 presents the (absolute) performance of RDT ERT and the second-best model per data set, which we choose based on averaged profit across deciles. Table 11 further displays the relative profit increase of RDT ERT compared to the respective runner-up. Results are again averages across data partitions. We highlight the highest profit values per decile in bold font. Interested readers find an extended version of Table 11 in the online appendix (see Table A3.2) where we report incremental campaign profit for all causal machine learning algorithms.

**Table 11: Incremental campaign profit of causal machine learning algorithms**

| Data | Causal machine learning algorithm | Incremental Campaign Profit Per Decile (€) | | | | | | | | | |
|---|---|---|---|---|---|---|---|---|---|---|---|
| | | Dec. 1 | Dec. 2 | Dec. 3 | Dec. 4 | Dec. 5 | Dec. 6 | Dec. 7 | Dec. 8 | Dec. 9 | Dec. 10 |
| **BAT** | RDT ERT | **5,668** | **8,109** | **9,115** | **10,202** | **11,275** | **12,266** | **13,262** | **14,144** | **14,896** | 15,135 |
| | X-Learner RF | 3,100 | 5,130 | 6,271 | 7,375 | 8,630 | 9,615 | 10,831 | 11,577 | 12,956 | 15,135 |
| | *Profit increase of RDT ERT (%)* | *82.8%* | *58.1%* | *45.4%* | *38.3%* | *30.6%* | *27.6%* | *22.4%* | *22.2%* | *15.0%* | *0.0%* |
| **FA** | RDT ERT | 548 | **938** | 1,161 | **1,443** | 1,553 | **1,739** | **1,985** | **2,143** | **2,301** | 2,400 |
| | Causal BART | **751** | 924 | **1,166** | 1,387 | **1,564** | 1,728 | 1,808 | 2,038 | 2,140 | 2,400 |
| | *Profit increase of RDT ERT (%)* | *-27.0%* | *1.5%* | *-0.4%* | *4.0%* | *-0.7%* | *0.6%* | *9.8%* | *5.2%* | *7.5%* | *0.0%* |
| **FB** | RDT ERT | 1,223 | 2,452 | **3,414** | **4,150** | **4,837** | **5,401** | **5,856** | **6,223** | **6,383** | 6,683 |
| | Causal Forest | **1,484** | **2,719** | 3,184 | 3,896 | 4,352 | 4,900 | 5,471 | 5,976 | 6,304 | 6,683 |
| | *Profit increase of RDT ERT (%)* | *-17.6%* | *-9.8%* | *7.2%* | *6.5%* | *11.1%* | *10.2%* | *7.0%* | *4.1%* | *1.3%* | *0.0%* |

Table 11 showcases the incremental campaign profit for each targeted decile. To increase the comprehensibility of a model's financial impact, we pick the first value of RDT from the BAT data and interpret it subsequently in detail. Therefore, using RDT would lead to a profit increase of 5,668€. From an operational perspective, a marketer would target the top ten percent of highest scored customers by the proposed revenue strategy, which would be 300 customers. We arrive at this number by considering the sampled customer population per data set (i.e., 100,000 observations) and dividing them by the number of ten data partitions. We further cut each partition into a 30% test fold and, given this example, target only customers from the first decile. To this end, based on the respective preceding campaign, RDT would target 300 customers, each of them providing an incremental profit of 18.89€ on average. Recall that in terms of our e-couponing settings, treatment costs are already recognized.

Recall that our data comes from real-word e-couponing campaigns. Considering the example of BAT, the actual campaign provided an incremental client revenue, or revenue uplift, of 9.76€ per person (see Table A3.1). More specifically, the average revenue per shop visitor has been 14.24€ and 4.48€ from the treatment and control group, respectively. Regarding the incentive costs of ten percent off a buyer's average purchase volume, we arrive at an incremental revenue of treated customers at 12.82€, while the average spend per customer from the control group has been 4.48€. Consequently, the incremental campaign profit per person of the actual BAT campaign was 8.34€. Our analysis suggests that targeting e-coupons using RDT substantially increases profit by 10.55€ per customer, which equates to a relative increase of 126.5% over the targeting model used for the actual campaign. The fact that the total incremental profit of 5,668€ (see Table 11) is an average across predictions on ten disjoint data partitions indicates that the result is robust.

We ascertain that incremental campaign profit differs across data sets. Whereas profit by targeting the whole customer population is 15,135€ and 6,683€ in terms of the BAT and FB data sets, respectively, it is comparably lower with 2,400€ in terms of the FA data set. We argue that this is due to the lower revenue per person and revenue uplift on the FA data set and refer to the varying customer spending behavior. The lower revenue per person from the treatment group is an essential aspect regarding its positive relation with campaign profit for uplift. Contrary to the FA data, the BAT and FB data have a 2.8 and 1.8 higher number of treated customers, and a 2.7 and 1.9 higher initial revenue uplift, respectively. The FA data might also have a higher degree of structural complexities, which complicates the identification of behavioral patterns.

Previous results evidence RDT as a promising strategy. However, we also observe sizeable profit increases over the actual campaign from the causal machine learning algorithms. More specifically, we identify the x-learner random forest, causal BART and causal forest as competitive models for the BAT, FA and FB data sets, respectively. The x-learner achieves a 24% higher incremental profit per person than the recent campaign for the first decile of the BAT data. Across targeted deciles of the FA data set, there is a tough competition between RDT and causal BART, which is reflected by the small difference of incremental profits for several deciles. Apart from this, causal BART considerably outperforms the proposed strategy with a 27% higher incremental profit on the first decile. Concerning the FB data, we draw a similar picture. To this end, the causal forest achieves a 17.6% and 9.8% better performance than RDT on the first and second deciles, respectively.

Overall, Table 11 re-emphasizes previous empirical results in favor of RDT, which provides the overall largest profit on most of the targeted deciles for each of the data sets. Averaging over deciles, RDT is 34.2%, 0.05%, and 2,0% better than the runner-up for the BAT, FA and FB data sets, respectively. Its strong financial achievements are particularly prevalent on BAT, where its relative profit increase ranges between 15.0% (ninth decile) and 82.2% (first decile). In this regard, RDT becomes even more important for operational campaign management as the largest relative difference of 82.2% relates to targeting the smallest fraction of customers from the whole population.

# 8 Summary, Implications and Limitations

Much prior work used conversion uplift models for targeting marketing campaigns. We introduced new target variable transformations to enable revenue uplift modeling. Assuming a marketing analyst to aim at maximizing the profit of targeted marketing activities, the proposed revenue uplift models offer a more direct and natural way to pursue this goal. Unlike conversion models, they account for heterogeneity in customer spending and target customers to maximize incremental revenue. Several empirical experiments demonstrated the effectiveness of the new RDT transformation. Across large amounts of real-world e-commerce data, it performed consistently better than alternative strategies for response, conversion uplift, and revenue uplift modeling in terms of business-oriented performance metrics. Comparisons to powerful causal machine learning algorithms further support this view. Examining the business impact of observed performance differences using a new proposed profit decomposition, we found RDT to provide sizeable improvements in profit compared to causal machine learning in an e-couponing context.

Empirical results have several implications for academia and corporate practice. From an academic perspective, the proposed target variable transformations extend existing approaches to develop causal uplift models to continuous responses. Causal inference is relevant in many disciplines. While the paper concentrates on targeting decisions in marketing, the RDT transformation could be used in other scientific domains to process continuous responses. Medical applications might be a good example. Examining the differential impact of a treatment (e.g., a new medication plan) on a continuous outcome variable (e.g., recovery time) is an exemplary application setting. Such settings are well studied in the literature and typically approached using causal models for treatment effect estimation. Our comparison to cutting-edge causal machine learning algorithms indicates that uplift transformations like RDT could extend the set of modeling tools in medical and other domains in a valuable way. A related point concerns the distribution of responses, which might introduce modeling challenges when transiting from discrete to continuous outcome variables. We have shown a possible approach to address zero-inflated response distributions using two-stage models. Our results in a marketing context do not evidence two-stage models to be useful. Whether modeling tasks in other domains lead to the same conclusion is a question for future research. In general, the flexibility to implement a causal model using any type of algorithm for regression/classification or more specialized techniques such as hurdle models is an important advantage of the uplift transformation approach over causal machine learning.

From a practitioner's point of view, new ways (i.e., RDT) to solve known problems (i.e., campaign planning) may also have some value. More importantly, the delineation between conversion and revenue uplift modeling is, to the best of our knowledge, originally proposed here. Marketing campaigns differ in their objective and may aim at lead generation, market growth, profit maximization, to name only a few. Uplift models are well-established in corporate practice. Our paper raises awareness for the point that different marketing objectives such as revenue/profit maximization are easier to accommodate in a model that predicts a continuous outcome variable. We also suggest and empirically compare concrete options to implement this concept using the proposed revenue uplift transformation or causal machine learning. We consider the algorithmic flexibility of the former a major advantage for marketing practice since it builds upon ex-isting technology. Considering the case of the RDT transformation, any software package capable of running logistic regression can be employed to build a revenue uplift model. Of course, causal machine learning is a viable alternative, and recent developments in the field such as causal forests enjoy much recognition. Our paper is one of the first to test corresponding techniques in a real-world marketing setting. While much more evidence in e-commerce and other marketing applications will be needed to have trust in their effectiveness in marketing (or lack thereof). Our results make a first step in this direction and may offer some guidance for research and development initiatives aiming at exploring causal machine learning in industry. Independent of a specific modeling method, practitioners may also benefit from the proposed profit decomposition. We design it as a new measure that better captures business goals than existing performance indicators for uplift modeling. The decomposition supports different cost models and thus campaign types. Measuring the incremental profit impact of a causal model, it bridges the gap between results from a data analytic model and financial implications in an easily interpretable manner.

Needless to say, our study has several limitations that open ways for future research. First, the employed marketing data set does not originate from a proper randomized trial, so that we cannot rule out the effect of confounders. For example, spill-over effects from parallel campaigns may have affected the customer behavior that we observe in the data. However, we deem such risk as low because the online marketing agency that supplied the data only displays one marketing incentive per customer session for the periods over which the data for this study was gathered. Consequently, for all recorded sessions in our data set, we can be sure that the marketing agency has not enacted other campaigns in parallel. Still, as the marketing agency acts in a consulting role to different online shops, there is a chance that the shops themselves have run further campaigns or commissioned further service providers. While many of the shops in our data set are relatively small and may not have the resources to carry out advanced campaigns themselves, we acknowledge this possible shortcoming. Furthermore, we employ data in a cross-sectional manner. Real-world marketing campaigns are often executed over a longer time and possibly with some adjustments. The cross-sectional setup may thus appear unrealistic. We strongly support the use of longitudinal data for an evaluation of (revenue) uplift modeling in future research, which was impossible here. Many of the shoppers in our data are new customers – or recurring customers that were not identified as such, due to, e.g., a lack of cross-device tracking – considering a panel setup would substantially reduce the amount of data available for analysis. Moreover, given that we obtain data from several different online shops over a relatively small period, a longitudinal setup appears less suitable. Although we consider several online shops with varying product categories and different times of data collection, a final limitation comes from the fact that all campaigns are based on e-coupons. Consequently, a replication of our study for other marketing applications and channels is highly desirable and an important task for future research.

# Acknowledgment

We are grateful to Fabian Gebert and colleagues from Akanoo for organizing access to the data for the empirical study. Valuable help with the empirical experiments from Tillmann Radmer is also acknowledged and much appreciated. We also thank four anonymous reviewers who provided several insightful comments to improve earlier versions of the paper. Finally, we very much appreciate the time and efforts of the editor, Prof. Robert Graham Dyson in handling our paper.

# Response Transformation and Profit Decomposition for Revenue Uplift Modeling

## – *Online Appendix* –

The online appendix includes supplementary material accompanying the main paper. It provides readers with a list of abbreviations and further results from empirical experiments. To this end, we provide information on the employed real-world data, learning algorithms, and performance measures (Section 5, "A1") as well as on further empirical analyses of uplift transformations for revenue uplift models (Section 6, "A2") and causal machine learning models (Section 7, "A3").

# List of Abbreviations

| | |
|---|---|
| BART | Bayesian Additive Regression Trees |
| BAT | Books and Toys |
| CATE | Conditional Average Treatment Effect |
| CRVTW | Continuous Response Variable Transformation with Weightings |
| DIY | Do-It-Yourself |
| ERT | Extremely Randomized Trees |
| FA | Fashion A |
| FB | Fashion B |
| GBC | Gradient Boosting Classifier |
| INDIRECT | Indirect Revenue Uplift Strategy |
| INDIRECT TS | Indirect Revenue Uplift Strategy Two-Stage Model |
| ITM | Interaction Term Method |
| KDD | Knowledge Discovery in Databases |
| KNN | K-Nearest Neighbor |
| LDA | Linear Discriminant Analysis |
| LDA WS | Linear Discriminant Analysis With Shrinkage |
| LL | Lasso LARS |
| LogR | Logistic Regression |
| LWUM | Lai's Weighted Uplift Method |
| MLP | Multi-Layer Perceptron Regressor |
| MM | Multimedia |
| OLS | Ordinary Least Squares |
| QDA | Quadratic Discriminant Analysis |
| RDT | Revenue Discretization Transformation |
| RESPONSE | Response Modeling |
| RFC | Random Forest Classifier |
| RFR | Random Forest Regressor |
| Ridge | Ridge Regression |
| SML | Supervised Machine Learning |
| SMOTE | Synthetic Minority Oversampling Technique |
| SVC | Support Vector Classifier |
| SVR | Support Vector Regressor |
| TCIA | Treatment-Covariates Interactions Approach |
| TSR | Theil-Sen Regressor |

# A1

# Further Details on Section 5 "Experimental Setup"

**Subsection A1.1: Empirical Data and Algorithm Configuration**

This section elaborates on the real-world e-commerce data that we employ in the main body of the paper. We first characterize the individual online shops that the original data set includes. Thereafter, we provide details on the independent variables that enable a prediction of shopping basket values. We produce these predictions using a set of machine learning algorithms for classification and regression. The last part of this section documents the hyperparameter candidate settings, which we consider when tuning individual learning algorithms.

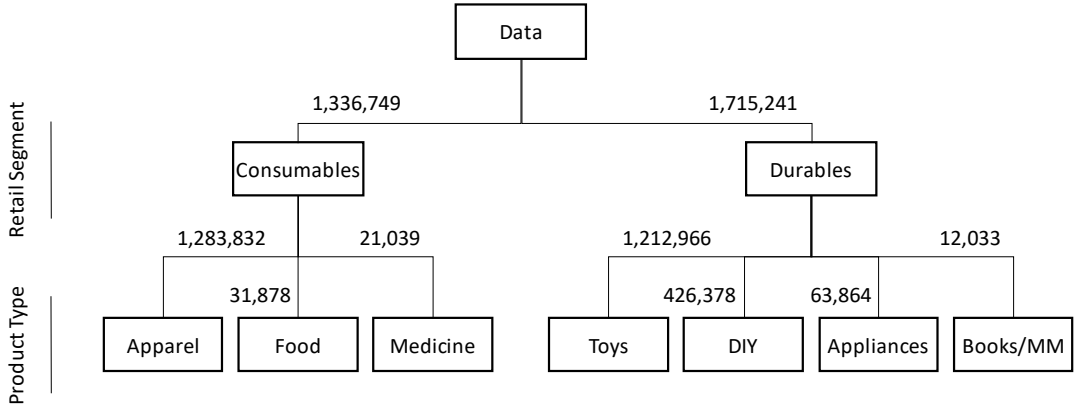

**Figure A1.1: Description of e-commerce data**

The shops operate exclusively in European national markets. The data includes shops from Austria, the Czech Republic, France, Germany, the Netherlands, Poland, and the United Kingdom. Figure A1.1 categorizes the data along with retail segments and product types; with information on the number of observations per category. Consumables are products of either single usage (such as food or medicine) or with a rather short lifecycle (e.g., apparel), and durables are low-wear products that are used multiple times over a longer time horizon. Related examples include "do-it-yourself" (DIY) products, toys, household appliances, books, and multimedia (MM).

**Table A1.1: Overview on independent variables**

| Variable Category | Variable Name | Description |
|---|---|---|
| Time | TimeToFirst (pageType)<br>TimeSinceFirst (pageType)<br>TimeSinceOn (pageType)<br>TimeOn (pageType)<br>SessionTime<br>HourOfDay | Time-related variables quantify customer preferences based on how long it takes, how much time is spent, etc. to visit a page of a certain page type. As "pageType" refers to CART, OVERVIEW, PRODUCT, SALE and SEARCH pages, TimeToFirst, TimeSinceFirst, TimeSinceOn and TimeOn variables consider all these different page types. Hence, this category lists these twenty variables plus total session time and the hour of the day the customer has entered the respective online shops. |
| Views | ViewCount<br>ViewedBefore (pageType=CART)<br>ViewsOn (pageType)<br>InitPageWas (pageType=OVERVIEW)<br>InitPageWas (pageType=PRODUCT)<br>InitPageWas (pageType=SALE)<br>NumDiffPages (pageType=OVERVIEW)<br>NumDiffPages (pageType=PRODUCT) | Views-related variables count the number of views per person; whether before swapping to another page type this person has accessed the shopping cart; the views on each page type other than CART (i.e., OVERVIEW, PRODUCT, SALE, SEARCH); and the initial page that the customer has accessed. Furthermore, the number of different overview- and product-related pages that the customer has visited is counted. |
| Basket | InitBasketNonEmpty<br>Normalized Basket Sum<br>Basket Quantity<br>HadBasketAdd<br>TimeToBasketAdd | Basket-related variables collect information on the basket's state at different points of time, such as whether the initial basket has been empty or not, how long it took until customers added products to their shopping baskets, the number of products at different states and the normalized basket sum to facilitate shop-wide comparisons of model performance. |
| Past Visit(s) | VisitorKnown<br>WasConvertedBefore<br>TimeSinceLastConversion<br>PreviousVisitCount<br>ViewCountLastVisit<br>TimeSinceLastVisit<br>DurationLastVisit<br>VisitCountLastWeek<br>VisitCountToday<br>TimeSinceFirstVisit | Variables that relate to past visits collect information such as whether the person is known or has purchased a certain product in the past and if so, how much time has passed since the last purchase. Furthermore, information is stored on how often the person re-visits the online shop within different periods, how much time has passed since the first visit (and independent from this, last visit) and the duration of the last visit. |
| Technicals | ScrollHeight (pageType=OVERVIEW)<br>ScreenWidth<br>Clicks (pageType=PRODUCT)<br>TimeSinceClick<br>TabSwitch (pageType=PRODUCT)<br>TimeSinceTabSwitch | Technical variables measure technology-related customer interactions for different page types. Thus, related variables identify the scroll height during the visit of overview-related pages, the screen width of the customer's respective device, how often customers switch browser tabs and the number of clicks on specific pages. |
| Meta-Variables | Conversion, Normalized Revenue<br>Shop-ID, Timestamp<br>Group Affiliation | Meta-variables identify if customers have converted in the current session; if so, how much (normalized) revenue they have generated; whether they have received an e-coupon; and which shop has been entered at what exact time. |

The data contains time-related variables such as the time passed on a certain page type. Views-related variables count the number of views for different page types. Basket-related variables collect information on shopping basket dynamics (e.g., basket quantities). Variables related to past customer visits and possible purchases are also available in our data. Technical variables focus on customer interactions with shop websites and a customer's related technical equipment. Meta-variables such as the shop identifier, timestamp, and group affiliation indicator further provide valuable information for our empirical study.

**Table A1.2: Learning algorithms and meta-parameter configuration for classification models**

| Learning Algorithm | Meta-Parameter Specification | Candidate Settings | Models |
|---|---|---|---|
| **Extremely Randomized Trees (ERT)** | Highest number of covariates randomly chosen for split<br>Minimum number of samples per leaf | 5, 15<br>200, 2000 | 4 |
| **Gradient Boosting Classifier (GBC)** | Maximal depth of regression estimators<br>Learning rate<br>Maximal number of features for split | 1, 3, 6<br>0.001, 0.01, …, 1<br>8, 12 | 24 |
| **k-Nearest-Neighbor (KNN)** | Number of nearest neighbors | 1, 5, 10, 100, 150, …, 500, 1000, …, 4000 | 19 |
| **Linear Discriminant Analysis with Shrinkage (LDA WS)** | Fixed shrinkage parameter | 0.05, …, 0.95 | 19 |
| **Logistic Regression (LogR)** | Regularization term<br>Regularization factor | L1, L2<br>1e-8, 1e-7, …, 1e8 | 34 |
| **Quadratic Discriminant Analysis (QDA)** | Regularization parameter for the covariance estimate | 0.10, …, 0.90 | 9 |
| **Random Forest Classifier (RFC)** | Highest number of covariates randomly chosen for split<br>Minimum number of samples per split | 8, 9<br>1000, 2000 | 4 |
| **Support Vector Classifier (SVC)** | Regularization factor<br>Calibration Method | 1e-10, 1e-9, …, 1e10<br>Sigmoid, Isotonic | 42 |
| | *Total number of classification models for RDT and for each conversion modeling strategy:* | | **155** |

**Table A1.3: Learning algorithms and meta-parameter configuration for regression models**

| Learning Algorithm | Meta-Parameter Specification | Candidate Settings | Models |
|---|---|---|---|
| **Lasso LARS (LL)** | Regularization parameter ($\alpha$) | 0.01, 0.1, …, 100 | 5 |
| **Multi-Layer Perceptron Regressor (MLP)** | Regularization parameter ($\alpha$)<br>Hidden layer sizes<br>Activation function for hidden layer<br>Learning rate<br>Initial learning rate for weights' updates | 0.01, 0.1, 1.0<br>(100, ), (100,100 ), (100,100,100 )<br>Rectified linear unit function<br>Invscaling<br>0.001, 0.01 | 18 |
| **Ridge Regression (Ridge)** | Regularization strength ($\alpha$) | 0.001, 0.01, …, 1000 | 7 |
| **Random Forest Regressor (RFR)** | Highest number of covariates randomly chosen for split<br>Minimum number of samples per leaf | 9, 12<br>100, 1000 | 4 |
| **Support Vector Regressor (SVR)** | Penalty parameter of the error term<br>Loss function | 1e-7, 1e-6, …, 1e4<br>(Squared) epsilon insensitive | 24 |
| **Theil-Sen Regressor (TSR)** | Maximal stochastic subpopulation | 10,000 | 1 |
| | *Total number of regression models for each revenue uplift strategy (except RDT):* | | **59** |

**Subsection A1.2: Qini-Related Performance Measures**

This section provides details on the performance measures that we apply to measure the effectiveness of a model and compare it to other models. To this end, we consider the Qini coefficient, which sums up model performance across deciles, and Qini curves that illustrate model performance on a decile-level.

The Qini coefficient and Qini curves are based on a rank ordering of treatment and control group observations according to their model-estimated uplift scores. For example, considering only the first ten percent of test set customers with highest uplift scores, the uplift of the model for the first decile is equal to the difference of the average response (i.e., the conversion rate for conversion uplift and customer spend for revenue uplift) among treatment and control group observations. Typically, effective uplift models allocate similar uplift scores to customers with similar behavioral patterns.

In addition to the property of incrementality, Revenue Qini curves are cumulative. Cumulativeness is important in a targeting context because targeting customers within higher deciles makes it necessary to target customers from preceding deciles as well (e.g., targeting customers in the fifth decile implies that the former 40% fraction of customers from the population are necessarily included). To obtain Revenue Qini curves, more technically, per decile and treatment and control group, we determine weighted revenue as we divide the revenue sum by the respective number of customers. We then take the difference between both groups and, to get a correct measure of total revenue, multiply the obtained weighted incremental revenue with the number of customers in the examined decile. Consequently, we obtain the incremental revenue sum for a specific targeted decile. The weighting is needed as the number of customers varies between treatment and control groups and targeted deciles. To calculate $Q_r$, we cumulate model performance along the targeted deciles. Furthermore, we create a linear line of random targeting that starts in (0,0) and ends in the model's incremental revenue sum from the tenth decile. Finally, for each decile, we subtract the cumulative value of random targeting from the model's respective incremental revenue sum to determine the area between the Revenue Qini curve and the random targeting line. Note that $Q_r$ values of zero indicate random performance. Negative $Q_r$ values imply worse model performance compared to random targeting with the model curve being on average below the random targeting line across deciles.

# A2

# Further Details on Section 6 "Empirical Analysis of Uplift Transformations for Revenue Uplift Models"

This section showcases results from further empirical analyses. More precisely, these include Revenue Qini curves of parametric revenue uplift models with one- and two-stage base learners, the variability of revenue uplift strategies with different SML algorithms for ten cross-validation iterations, and for the best algorithm per strategy, respective Revenue Qini curves that exhibit financial returns from targeting.

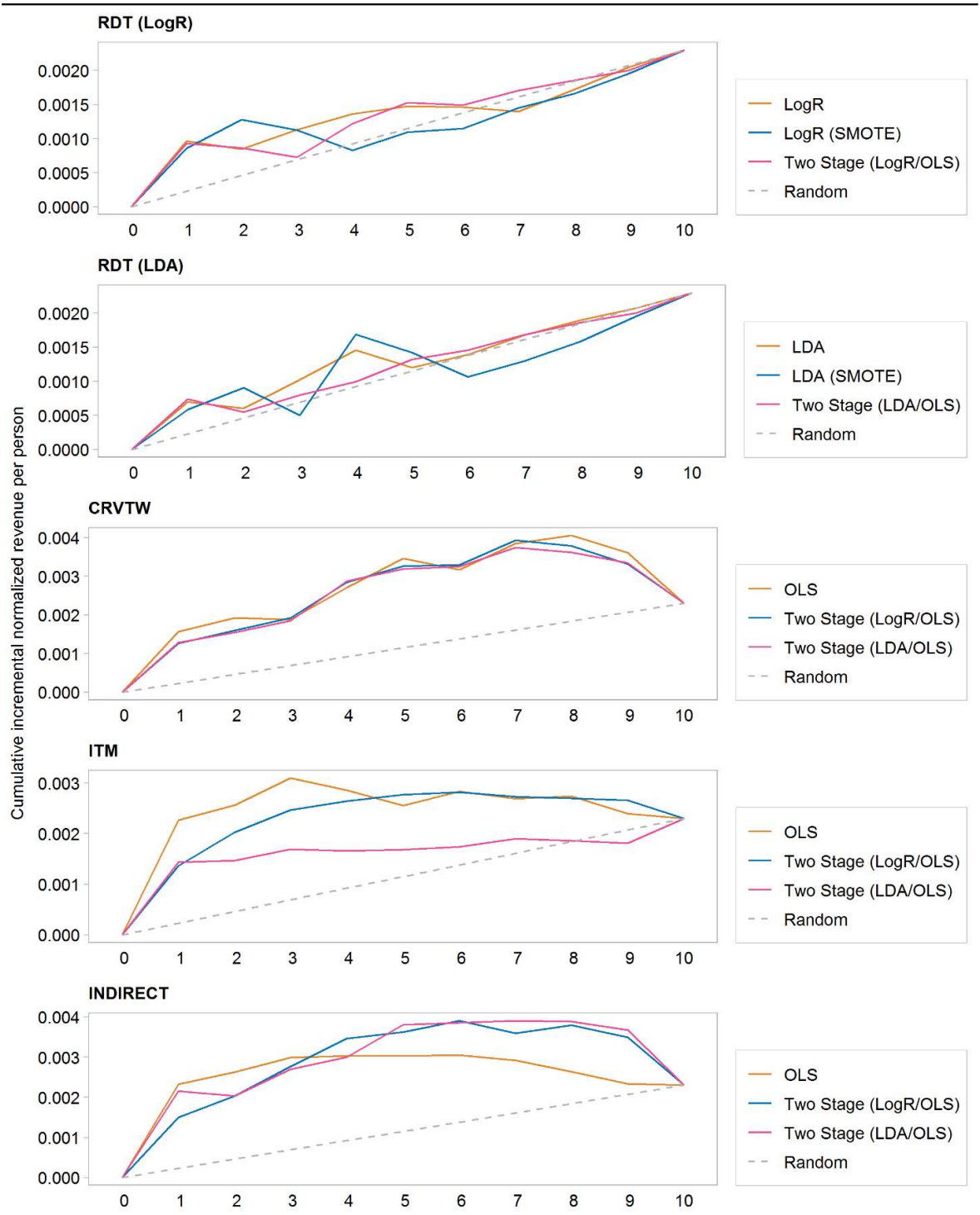

**Figure A2.1: Revenue Qini curves of revenue uplift strategies with parametric models**

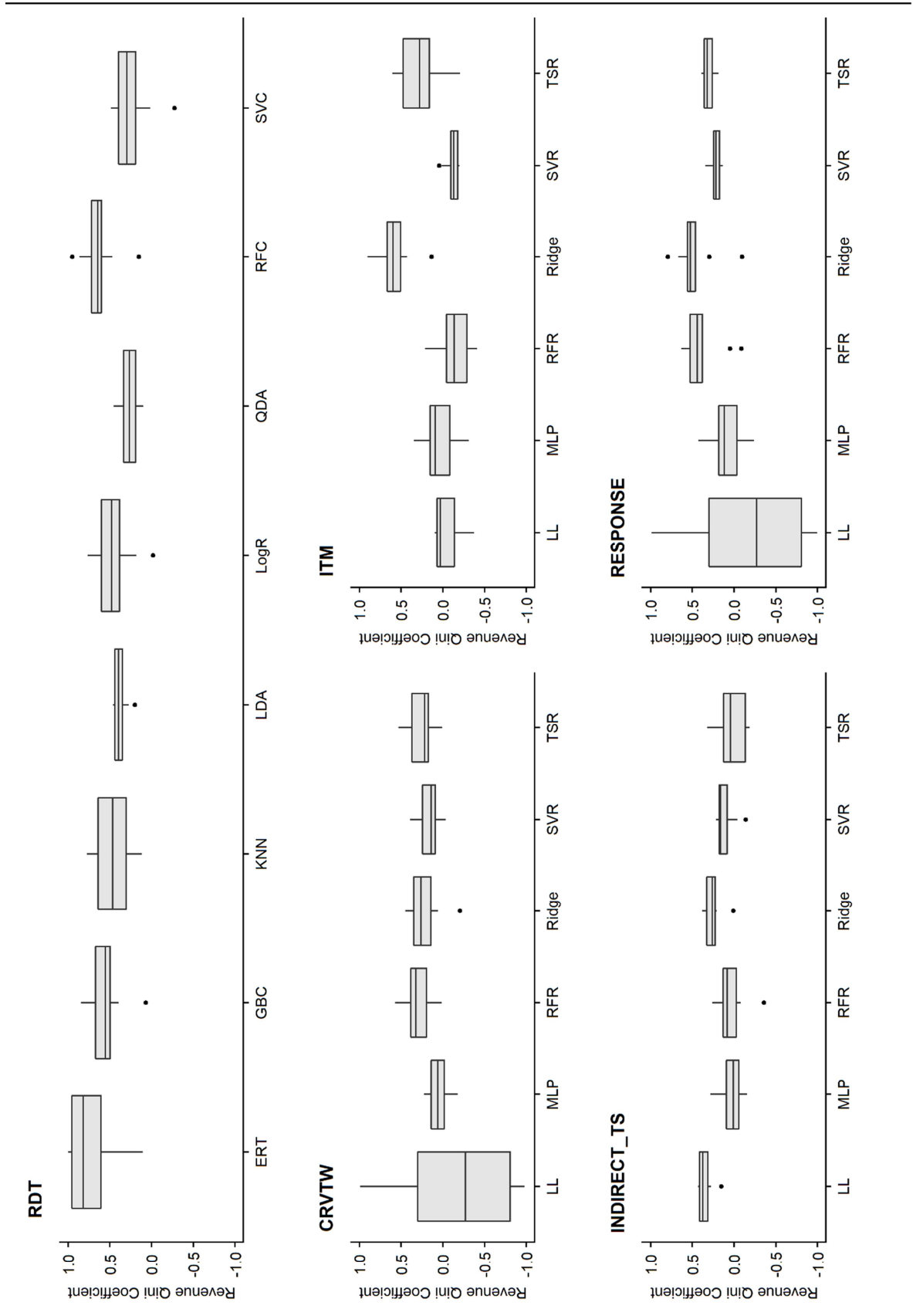

**Figure A2.2: Distribution of revenue models over ten cross-validation iterations**

We summarize our findings from Figure A2.1 and Figure A2.2 as follows. In short, Figure A2.1 provides three key insights. First, we see an almost steady increase of model performance for CRVTW, ITM and INDIRECT TS on the first targeted deciles. Second, model performance greatly differs between the one-stage and two-stage models for each of the revenue uplift strategies. Third, Figure A2.1 confirms the model performance results in terms of $Q_r$.

Figure A2.2 demonstrates that RDT achieves higher uplift and less variation across cross-validation iterations compared to the other revenue uplift strategies. Except for the Lasso LARS model, we also observe appealing results in the form of low variation and relatively high uplift for CRVTW which represents the second proposed revenue uplift strategy and is instrumental for RDT. Furthermore, Figure A2.2 suggests that the ITM-based revenue uplift models and the indirect revenue uplift two-stage models predict less accurately than comparable competitors. However, the indirect approach with the focus on the zero-inflated revenue distribution overall provides the lowest variability among the regression-based revenue uplift strategies. An exception is related to Theil-Sen regression which, in the form of a response model, is most competitive in terms of the variability assessment. While the random forest regression model of CRVTW achieves a better performance than random forest regressions of other revenue uplift strategies, its variation is comparable. Among these revenue uplift models, the ITM-based random forest regression has the highest variability.

Considering supervised learning methods, by comparing the interquartile range across learners, the good performance of ridge regressions is not limited by a potential high variability in the results. This is valid for all regression-based revenue strategies in comparison with other underlying models. Particularly, score distributions from Lasso LARS regression models have a high mid-spread, which stresses the instability of their models' results. Given the high uncertainty from two out of four revenue uplift strategies, we refrain from further application of Lasso LARS revenue uplift models and instead qualify the second-best two-stage model, a ridge regression, for further analyses in this paper.

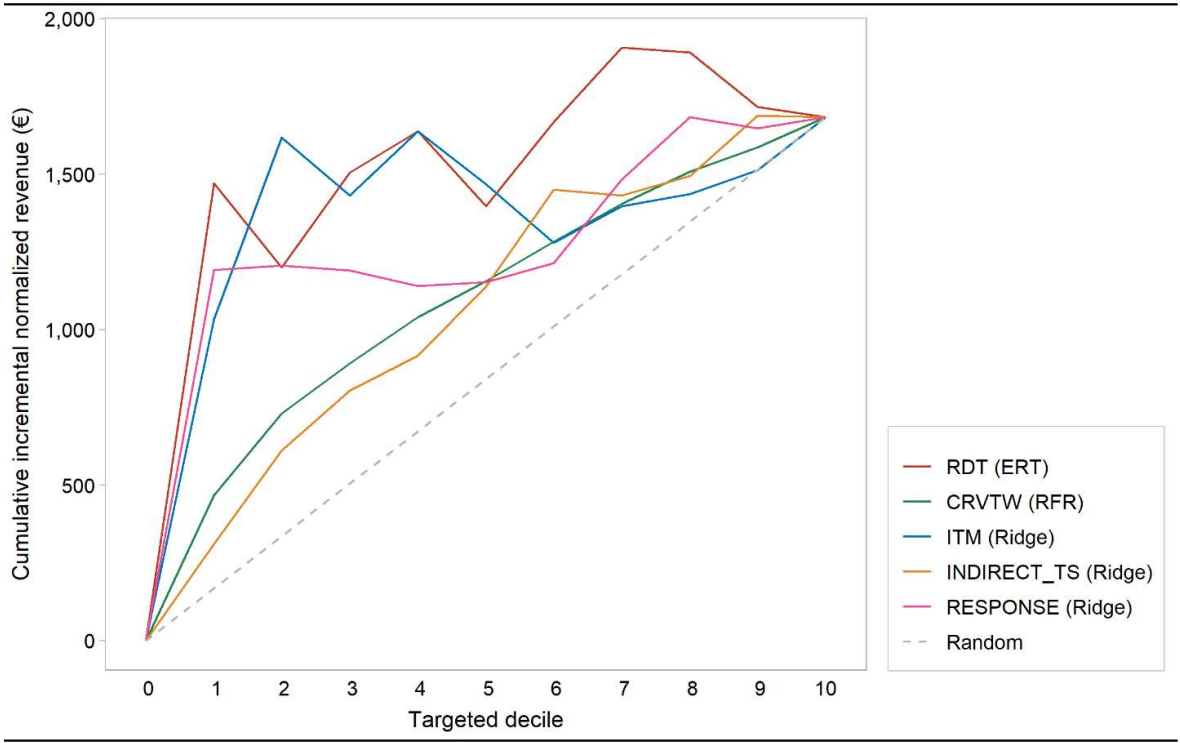

**Figure A2.3: Revenue Qini curves of selected revenue uplift strategies**

The revenue normalization allows comparisons with data from other experimental settings (e.g., different product categories or online shops). The horizontal axis exhibits the fraction of customers to receive a specific treatment. To gain relevance, a model must exceed a pre-defined general threshold which is displayed with a random, dotted targeting line.

All Revenue Qini curves start with an increase in the first decile. Whereas the RDT model is superior when targeting 10%, 30%, 60%, 70%, 80%, and 90% of customers, ITM dominates on the 20%, 40%, and 50% fractions of the total customer base. The remaining strategies are not competitive in terms of the examined financial performance measure. Targeting 100% of customers (i.e., the whole customer population) equals the effectiveness of random targeting and does not provide more value than model-based targeting.

Except on the first and eighth deciles where the response modeling-based ridge regression achieves second-best performance, it does not show good results across deciles. Although much better than random targeting on each targeted decile, CRVTW and the two-stage model on the indirect approach do not deliver promising results and are even outperformed by the revenue response model except on the fifth and sixth deciles. Especially on the first five deciles, CRVTW and the indirect approach have substantial value differences to RDT and ITM. However, despite its inferior model performance, the Qini curve of CRVTW is inherently smooth and increases on every decile.

# A3
# Further Details on Section 7 "Analysis of RDT Uplift Strategy Against Causal Machine Learning"

**Subsection A3.1: Implementation and Specification of Causal Machine Learning Algorithms**

In this section, we focus on the available key software packages, our hardware and the setup of the examined causal machine learning algorithms. Subsequently, we list the employed software packages, some of them which we have downloaded from Github repositories. These are "extraTrees" (RDT ERT), "susanathey/causalTree" (causal tree), "grf" (causal forest), "saberpowers/causalLearning" (causal boosting), "vdorie/bartCause" (causal BART), and "xnie/rlearner" (x-learner). To conduct the analysis, we use a terminal server with 288 gigabytes random-access memory (RAM) and twenty-four cores at 3.4 gigahertz.

Furthermore, to provide a fair assessment of model performance, we try to specify the algorithms as equal as possible. Therefore, we disregard using multiple processing units as only some models offer a related hyperparameter specification (e.g., ERT or causal forest). More specifically, we build 500 trees on each of the forest-based models (i.e., RDT ERT, causal forest, causal boosting, and the x-learner random forest). For the RDT ERT model, causal forest and x-learner random forest, we consider five randomly sampled covariates for each tree split. The RDT ERT model further takes five random cuts as per random covariate into account. For the causal tree, we consider sub-sample or honest splitting and use the honest version of the causal tree splitting rule. Hence, we separate model building from prediction by using different samples (Athey & Imbens, 2016). We set the scale parameter to 0.5 to calculate a rather conservative honest risk evaluation function for cross-validation. We further use a minimum of 50 treated and 50 control observations at each leaf. As with the causal tree model, we use honest splitting for the causal forest model and thereby imitate experimental setups of other recent studies (e.g., Knaus et al., 2018). Also, we keep at least ten cases in every leaf of the forest and grow two trees per subsample. Due to the empirical merit of further considering local centering for causal forests (e.g., Athey et al., 2019), we also specify the causal forest models for our study accordingly. Finally, we use orthogonalization to predict the estimates of the response and treatment variables utilizing boosted regression forests.

We add a causal boosting model to our experimental setting with default parameters. More specifically, this model is made up of 500 shallow causal trees with at most four tree leaves and considers a learning rate of 0.01. The distance between tree splits is set to 0.1. The causal BART model fits response and treatment models to determine treatment effects. In particular, we fit the response surface (i.e., the response given treatment and covariates, see Hill (2011)) with BART to determine CATE estimates $\widehat{Y}_i(1)$ and $\widehat{Y}_i(0)$ and average the differences between these estimates. Next, we fit a separate BART algorithm to the treatment variable given the confounders. As a result of model predictions, we get CATE observations across the number of samples and Markov chains. To obtain a vector of CATE estimates with an equal size as the number of test set observations, we average these results accordingly. As indicated above, for the x-learner random forest, we only set the number of trees to 500 and use the function's default parameters for regression. The x-learner strategy provides a framework to integrate arbitrary supervised machine learning algorithms. Our choice of the underlying random forest model is further motivated by related literature from medical trials (Duan et al., 2019).

## Subsection A3.2: Additional E-Commerce Data Sets

The comparison of the proposed RDT strategy for revenue uplift modeling against recently developed causal machine learning algorithms in the main body of the paper employs auxiliary data. This section elaborates on corresponding e-commerce data sets, which refer to three online shops in which couponing campaigns have been executed. We first report summary statistics of the campaign uplift and subsequently discuss independent variables.

**Table A3.1: Descriptive uplift statistics of out-of-universe data**

| Data | Group Affiliation | Share [%] | Customer Sessions | Purchasers | Conversion Rate [%] | Conversion Uplift [%] | Revenue [€] | Revenue Per Person [€] | Revenue Uplift [€] |
|---|---|---|---|---|---|---|---|---|---|
| BAT | Treatment | 74.8 | 111,729 | 17,890 | 16.01 | 0.72 | 1,591,459 | 14.24 | 9.76 |
|  | Control | 25.2 | 37,570 | 5,745 | 15.29 |  | 500,579 | 4.48 |  |
|  | Total | 100 | 149,299 | 23,635 | - | - | 2,092,039 | - | - |
| FA | Treatment | 75.0 | 861,763 | 44,045 | 5.11 | 0.47 | 4,393,086 | 5.10 | 3.56 |
|  | Control | 25.0 | 287,016 | 13,311 | 4.64 |  | 1,322,738 | 1.53 |  |
|  | Total | 100 | 1,148,779 | 57,356 | - | - | 5,715,823 | - | - |
| FB | Treatment | 75.0 | 501,366 | 47,024 | 9.38 | 0.45 | 4,824,085 | 9.62 | 6.60 |
|  | Control | 25.0 | 166,909 | 14,908 | 8.93 |  | 1,515,925 | 3.02 |  |
|  | Total | 100 | 668,275 | 61,932 | - | - | 6,340,010 | - | - |

We perform Pearson's chi-squared tests to check whether conversion uplift is statistically significant. We find that this is not the case for the BAT data (p-value of 9.34e-04), FA data (p-value smaller than 2.2e-16), and FB data (p-value of 4.78e-08). We conclude that the conversion uplift signals of these datasets are rather low although they are on average three times higher than on the original dataset. We emphasize that the BAT data has the highest revenue uplift among all data sets.

In contrast to the original dataset, the out-of-universe datasets are more focused towards specific products. To this end, the books and toys (BAT) data comprise 149,299 observations, the fashion A (FA) data consists of 1,148,779 observations and the fashion B (FB) data has 668,275 observations. Apart from the number of customer sessions per dataset, the out-of-universe data contains the same variables as the original data. Apart from them, the BAT, FA and FB data sets include 33 further variables that measure which channel the customer uses (i.e., e-mail, paid, or search), the number of seconds on dedicated pages, to which specific page type the current page belongs to, and some more meta-variables (e.g., whether a customer has aborted a session).

## Subsection A3.3: Additional Empirical Results

This section discusses the decile-specific results of causal machine learning models in terms of Revenue Qini curves and incremental campaign profit and further shows profit results of RDT ERT per data partition.

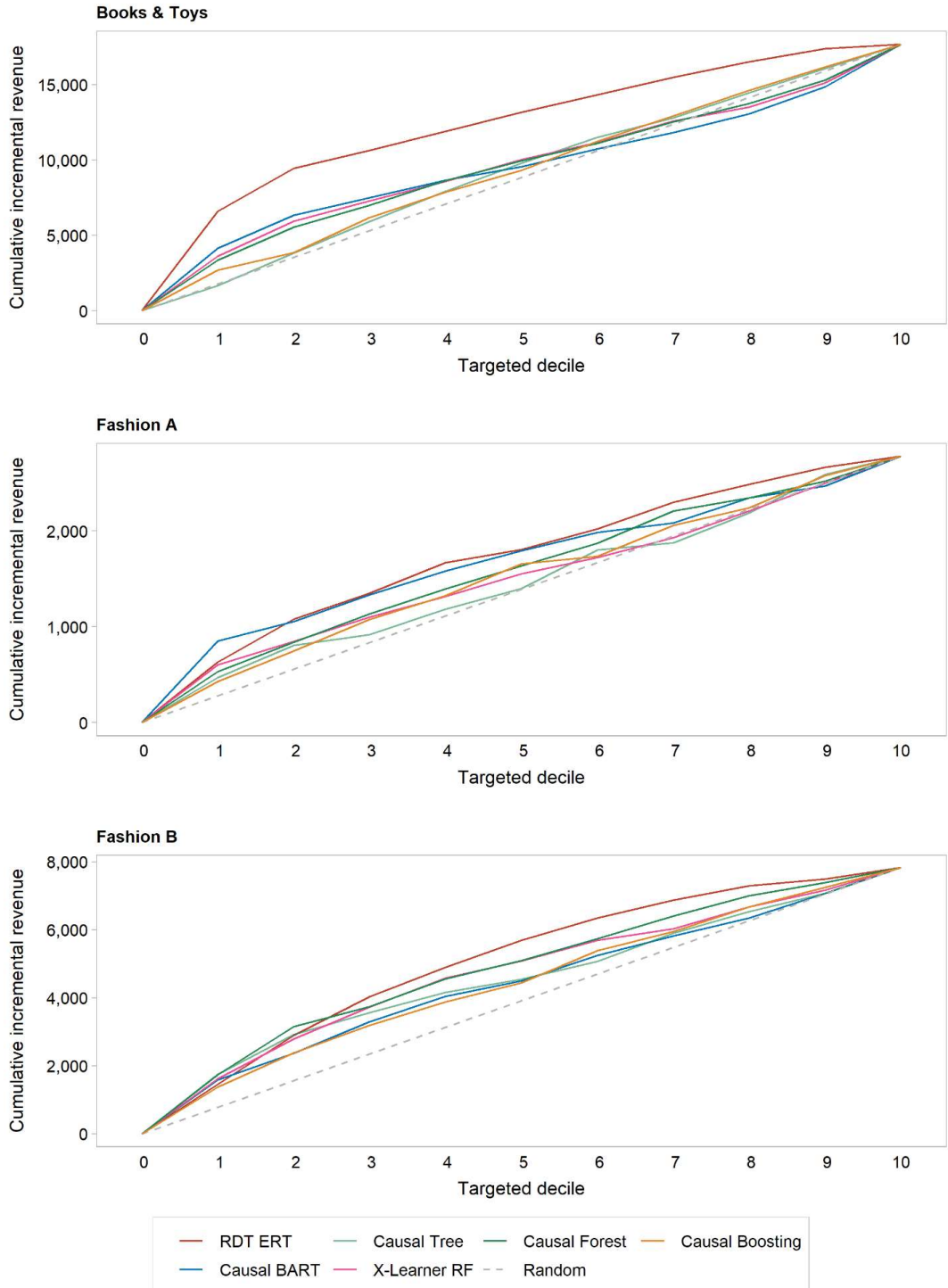

**Figure A3.1: Revenue Qini curves of causal machine learning models on different datasets**

**Table A3.2: Incremental campaign profit of causal machine learning algorithms**

| Data | Causal machine learning algorithm | Incremental Campaign Profit Per Decile (€) | | | | | | | | | |
|---|---|---|---|---|---|---|---|---|---|---|---|
| | | Dec. 1 | Dec. 2 | Dec. 3 | Dec. 4 | Dec. 5 | Dec. 6 | Dec. 7 | Dec. 8 | Dec. 9 | Dec. 10 |
| **BAT** | RDT ERT | **5,668** | **8,109** | **9,115** | **10,202** | **11,275** | **12,266** | **13,262** | **14,144** | **14,896** | 15,135 |
| | Causal Tree | 1,410 | 3,289 | 5,082 | 6,826 | 8,447 | *9,896* | 10,970 | 12,397 | 13,767 | 15,135 |
| | Causal Forest | 2,895 | 4,774 | 6,013 | 7,414 | 8,546 | 9,519 | 10,754 | 11,774 | 13,113 | 15,135 |
| | Causal Boosting | 2,330 | 3,304 | 5,301 | 6,761 | 7,973 | 9,644 | *11,084* | *12,511* | *13,856* | 15,135 |
| | Causal BART | *3,561* | *5,454* | *6,446* | *7,448* | 8,188 | 9,189 | 10,125 | 11,177 | 12,730 | 15,135 |
| | X-Learner RF | 3,100 | 5,130 | 6,271 | 7,375 | *8,630* | 9,615 | 10,831 | 11,577 | 12,956 | 15,135 |
| **FA** | RDT ERT | *548* | **938** | *1,161* | **1,443** | *1,553* | **1,739** | **1,985** | **2,143** | **2,301** | 2,400 |
| | Causal Tree | 411 | 706 | 795 | 1,023 | 1,211 | 1,570 | 1,619 | 1,886 | *2,244* | 2,400 |
| | Causal Forest | 465 | 734 | 991 | 1,216 | 1,424 | 1,630 | *1,929* | *2,041* | 2,185 | 2,400 |
| | Causal Boosting | 362 | 641 | 929 | 1,141 | 1,432 | 1,485 | 1,775 | 1,926 | 2,223 | 2,400 |
| | Causal BART | **751** | *924* | **1,166** | *1,387* | **1,564** | *1,728* | 1,808 | 2,038 | 2,140 | 2,400 |
| | X-Learner RF | 529 | 728 | 944 | 1,131 | 1,338 | 1,490 | 1,665 | 1,912 | 2,165 | 2,400 |
| **FB** | RDT ERT | 1,223 | 2,452 | **3,414** | **4,150** | **4,837** | **5,401** | **5,856** | **6,223** | **6,383** | 6,683 |
| | Causal Tree | *1,479* | *2,500* | 3,044 | 3,553 | 3,880 | 4,333 | 5,046 | 5,582 | 6,032 | 6,683 |
| | Causal Forest | **1,484** | **2,719** | 3,184 | 3,896 | 4,352 | *4,900* | *5,471* | *5,976* | *6,304* | 6,683 |
| | Causal Boosting | 1,178 | 2,040 | 2,725 | 3,312 | 3,775 | 4,602 | 5,074 | 5,702 | 6,188 | 6,683 |
| | Causal BART | 1,354 | 2,007 | 2,806 | 3,448 | 3,818 | 4,466 | 4,956 | 5,396 | 6,038 | 6,683 |
| | X-Learner RF | 1,373 | 2,387 | *3,188* | *3,932* | *4,354* | 4,876 | 5,161 | 5,718 | 6,130 | 6,683 |

We summarize our findings from Figure A3.1 and Table A3.2 as follows. From Figure A3.1 we identify RDT ERT, causal BART, causal forest, and the x-learner random forest as generally competitive. They outperform causal boosting and the causal tree. More specifically, RDT ERT is the overall best model in terms of cumulative incremental revenue for most of the targeted deciles. Its superiority is most pronounced on the BAT data, where it sets itself apart on the first decile and keeps its dramatic merit over all other models until the tenth decile. The causal BART algorithm is the second-best model across the datasets. For the first six deciles of the FA data, it achieves similar performance as RDT ERT and outperforms all remaining algorithms on these deciles. It is also the second most competitive model on the first three deciles on the BAT data, which have special importance for targeted decision-making as shown in the previous analysis. Interestingly, similar to the trend of the causal forest and x-learner random forest, the causal BART algorithm is not even as good as random targeting for the latter deciles on this dataset.

Apart from this, we see that the causal forest ranks third overall, followed by the x-learner random forest. On the first two deciles of the FB data, the causal forest achieves superior results than all remaining models. While its performance is slightly worse compared to the x-learner random forest on the BAT data, it is better on the fashion-specific datasets. This is particularly the case for the latter deciles. Although better than random targeting, we find that causal boosting is not competitive in terms of cumulative incremental revenue across datasets. Among all causal machine learning algorithms, the performance of the causal tree is overall lowest. The only performance peak we note is based on the first three deciles of the FB data, where the performance gap between the causal tree and the top three models is rather small. The results of the causal tree do not come as a surprise given that it only consists of a single tree.

In general, Table A3.2 further supports our findings. This does not come as a surprise although the table displays incremental campaign profit, and not incremental revenue, as the model ranking should not differ too much between the two measures of financial performance. We emphasize that the proposed revenue strategy outperforms its competitors on all deciles of the BAT data, and is further achieving the top performance on most of the targeted deciles regarding the FA and FB data sets.

**Table A3.3: Incremental campaign profit of RDT ERT for each of the ten data partitions**

| Data | Partition | Incremental Campaign Profit Per Decile (€) | | | | | | | | |
|---|---|---|---|---|---|---|---|---|---|---|
| | | Dec. 1 | Dec. 2 | Dec. 3 | Dec. 4 | Dec. 5 | Dec. 6 | Dec. 7 | Dec. 8 | Dec. 9 |
| BAT | 1 | 2,871 | 5,444 | 6,733 | 7,675 | 8,529 | 8871 | 10,019 | 11,474 | 11,761 |
| | 2 | 6,652 | 8,546 | 9,375 | 10,515 | 12,018 | 12,873 | 13,612 | 14,715 | 15,435 |
| | 3 | 4,744 | 7,755 | 7,471 | 8,714 | 10,647 | 11,796 | 12,680 | 13,523 | 14,644 |
| | 4 | 7,007 | 9,277 | 10,096 | 10,813 | 12,140 | 12,891 | 13,369 | 14,784 | 15,341 |
| | 5 | 6,241 | 7,240 | 8,512 | 9,448 | 10,475 | 12,274 | 13,927 | 14,513 | 15,131 |
| | 6 | 5,574 | 8,881 | 10,163 | 11,497 | 12,688 | 13,433 | 14,097 | 14,413 | 14,854 |
| | 7 | 4,809 | 7,407 | 8,844 | 10,293 | 11,023 | 12,621 | 14,184 | 15,482 | 16,178 |
| | 8 | 7,289 | 10,540 | 11,603 | 12,592 | 13,427 | 14,213 | 15,338 | 16,779 | 17,533 |
| | 9 | 5,546 | 7,333 | 8,103 | 9,336 | 10,317 | 11,160 | 12,051 | 12,091 | 13,422 |
| | 10 | 5,947 | 8,662 | 10,247 | 11,138 | 11,484 | 12,529 | 13,339 | 13,666 | 14,660 |
| | *Mean* | *5,668* | *8,109* | *9,115* | *10,202* | *11,275* | *12,266* | *13,262* | *14,144* | *14,896* |
| FA | 1 | -235 | 91 | 97 | 236 | 194 | 474 | 873 | 1,137 | 1,548 |
| | 2 | -43 | 266 | 792 | 852 | 815 | 1,129 | 12,49 | 1,680 | 2,036 |
| | 3 | -25 | 487 | 1,025 | 1,168 | 1,496 | 1,814 | 1,665 | 1,796 | 1,925 |
| | 4 | 1,792 | 2,231 | 2,645 | 3,270 | 3,647 | 3,647 | 3,852 | 3,852 | 3,852 |
| | 5 | 513 | 945 | 1,166 | 1,166 | 1,221 | 1,491 | 1,491 | 1,718 | 1,837 |
| | 6 | 363 | 1,080 | 1,312 | 1,842 | 1,708 | 1,820 | 1,959 | 1,816 | 1,778 |
| | 7 | 700 | 1,350 | 1,627 | 2,205 | 2,191 | 2,298 | 3,032 | 2,924 | 3,401 |
| | 8 | 1,055 | 1,303 | 1,337 | 1,799 | 1,548 | 1,666 | 2,087 | 2,535 | 2,535 |
| | 9 | 534 | 800 | 1,112 | 1,100 | 1,364 | 1,384 | 1,470 | 1,707 | 1,670 |
| | 10 | 822 | 823 | 498 | 795 | 1,350 | 1,668 | 2,169 | 2,266 | 2,423 |
| | *Mean* | *548* | *938* | *1,161* | *1,443* | *1,553* | *1,739* | *1,985* | *2,143* | *2,301* |
| FB | 1 | 1,211 | 2,423 | 3,513 | 4,312 | 4,725 | 5,862 | 6,208 | 6,402 | 6,960 |
| | 2 | 1,246 | 2,570 | 3,022 | 4,676 | 5,851 | 6,641 | 7,131 | 7,485 | 7,335 |
| | 3 | 1,262 | 3,030 | 4,309 | 5,311 | 5,726 | 5,702 | 6,260 | 6,389 | 6,346 |
| | 4 | 2,011 | 2,932 | 4,442 | 4,896 | 5,151 | 5,603 | 6,043 | 6,740 | 7,154 |
| | 5 | 1,388 | 2,478 | 3,957 | 4,219 | 5,303 | 5,748 | 6,047 | 6,561 | 6,730 |
| | 6 | 724 | 1,577 | 1,485 | 2,505 | 3,423 | 4,064 | 4,877 | 5,200 | 5,326 |
| | 7 | 293 | 1,286 | 2,896 | 3,463 | 4,187 | 5,137 | 5,935 | 5,938 | 6,048 |
| | 8 | 1,633 | 3,283 | 3,724 | 4,678 | 5,330 | 5,684 | 6,077 | 6,954 | 6,684 |
| | 9 | 1,304 | 2,236 | 3,384 | 3,899 | 4,236 | 4,437 | 4,621 | 4,932 | 5,277 |
| | 10 | 1,159 | 2,703 | 3,403 | 3,539 | 4,433 | 5,131 | 5,357 | 5,626 | 5,968 |
| | *Mean* | *1223* | *2,452* | *3,414* | *4,150* | *4,837* | *5,401* | *5,856* | *6,223* | *6,383* |

Table A3.3 shows incremental campaign profit of RDT ERT per data partition and out-of-universe data set.